\documentclass{article}
\usepackage[left=1.1 in,right=1.1 in,top=1.1 in,bottom=1.1 in]{geometry}
\usepackage{graphicx, subfigure}
\usepackage{amssymb}
\usepackage{amsfonts}
\usepackage{amsmath}
\usepackage{geometry}
\usepackage{lscape}
\usepackage{picins}
\usepackage{scalefnt}
\usepackage{appendix}
\usepackage{titlesec}
\usepackage{setspace}
\usepackage{fancyhdr}
\usepackage{mathrsfs}
\usepackage{multirow}

\usepackage{enumitem} % for noitemsep
\usepackage{url}

\usepackage[affil-it]{authblk}
\usepackage{algorithm}
\usepackage{algorithmic}

\newtheorem{theorem}{Theorem}

\newtheorem{lemma}[theorem]{Lemma}

\DeclareMathOperator*{\argmin}{arg\,min}

\allowdisplaybreaks[4]
\title{Algorithms for Generalized Cluster-wise Linear Regression}

\date{May 17, 2016}

\begin{document}

\author{Young Woong Park\thanks{ywpark@smu.edu, Cox School of Business, Southern Methodist University, Dallas, TX, USA.} \qquad Yan Jiang\thanks{jiangyan1984@gmail.com, Sears Holdings Corporation, Hoffman Estates, IL} \qquad Diego Klabjan\thanks{d-klabjan@northwestern.edu, Department of Industrial Engineering and Management Sciences, Northwestern University, Evanston, IL, USA.} \qquad Loren Williams\thanks{loren.williams@ey.com, Ernst \& Young LLP, Atlanta, GA} }

\maketitle

\begin{abstract}
Cluster-wise linear regression (CLR), a clustering problem intertwined with regression, is to find clusters of entities such that the overall sum of squared errors from regressions performed over these clusters is minimized, where each cluster may have different variances. We generalize the CLR problem by allowing each entity to have more than one observation, and refer to it as generalized CLR. We propose an exact mathematical programming based approach relying on column generation, a column generation based heuristic algorithm that clusters predefined groups of entities, a metaheuristic genetic algorithm with adapted Lloyd's algorithm for K-means clustering, a two-stage approach, and a modified algorithm of Sp{\"a}th \cite{Spath1979} for solving generalized CLR. We examine the performance of our algorithms on a stock keeping unit (SKU) clustering problem employed in forecasting halo and cannibalization effects in promotions using real-world retail data from a large supermarket chain. In the SKU clustering problem, the retailer needs to cluster SKUs based on their seasonal effects in response to promotions. The seasonal effects are the results of regressions with predictors being promotion mechanisms and seasonal dummies performed over clusters generated. We compare the performance of all proposed algorithms for the SKU problem with real-world and synthetic data.
\end{abstract}

\section{Introduction}
%Clustering is a commonly encountered problem in many areas such as marketing, engineering, biology, etc. 
Clustering is a commonly encountered problem in many areas such as marketing, engineering, and biology, among others. 
In a typical clustering problem, the goal is to group entities together according to a certain similarity measure. Such a measure can be defined in many different ways, and it determines the complexity of solving the relevant clustering problem. Clustering problems with the similarity measure defined by regression errors is especially challenging because it is coupled with regression.

Consider a retailer that needs to forecast sales at the stock keeping unit (SKU) level for different promotional plans and mechanisms (e.g., 30\% off the selling price) using a linear regression model. A SKU is a unique identifying number that refers to a specific item in inventory. Each SKU is often used to identify product, product size, product type, and the manufacturer. Seasonality is an important predictor and is modeled using an indicator dummy input variable for each season, with the length of one season being one week. The usable data for each SKU is limited compared to the possible number of parameters to estimate, among which the seasonality dummies compose a large proportion. More significant and useful statistical results can be obtained by clustering SKUs with similar seasonal effects from promotions together, and estimating seasonality dummies for a cluster instead of a single SKU. However, the seasonal effects of SKUs correspond to regression coefficients, which can only be obtained after grouping SKUs with similar seasonality.

A two-stage method can be used to solve such difficult clustering problems that are intertwined with regression. In the first stage, entities are clustered based on certain approximate measures of their regression coefficients. In the second stage, regressions are performed over the resultant clusters to obtain estimates for the regression coefficients for each cluster. However, good approximate measures are difficult to obtain a priori before carrying out the regressions. A better alternative is to perform clustering and regression simultaneously, which can be achieved through cluster-wise linear regression (CLR), which is also referred to as ``regression clustering'' in the literature. Other application areas of CLR include marketing, pavement condition prediction, and spatial modeling and analysis. More details about these other application areas can be found in Openshaw \cite{Openshaw1976}, DeSarbo and Cron \cite{DeSarbo1988}, DeSarbo \cite{DeSarbo1989}, and Luo and Chou \cite{Luo2006}.

The CLR problem bears connection to the minimum sum-of-squares clustering (MSSC) problem, the objective of which is to find clusters that minimize the sum of squared distances from each entity to the centroid of the cluster which it belongs to. Contrary to clustering entities directly based on distances, CLR generates clusters according to the effects that some independent variables have on the response variable of a preset regression model. Each entity is represented by a set of observations of a response variable and the associated predictors. CLR is to group entities with similar regression effects into a given number of clusters such that the overall sum of squared residuals within clusters is minimal. Although the MSSC problem has been extensively studied by researchers from various fields (e.g., statistics, optimization, and data mining), the work for the CLR problem is limited, most of which concerns adapting the Lloyd's algorithm based heuristic algorithms of the MSSC problem to the CLR problem. The Lloyd's algorithm starts randomly from some initial partition of clusters, then calculates the centroids of clusters, and assigns entities to their closest centroids until converging to a local minimum. Recently, several exact approaches have been proposed by Carbonneau et al \cite{Carbonneau2011,Carbonneau2012,Carbonneau2014}, which are discussed in detail in Sections 1.1 and 2.

We tackle the problem of clustering entities based on their regression coefficients by modeling it as a generalized CLR problem, in which we allow each entity to have more than one observation. We propose both a mixed integer quadratic program formulation and a set partitioning formulation for generalized CLR. Our mixed integer quadratic program formulation is more general than the one proposed by Bertsimas and Shioda \cite{Bertsimas2007}, which cannot be directly applied to the SKU clustering problem since they assume each clustering entity to have only one observation and this assumption does not hold for the SKU clustering problem. We identify a connection between the generalized CLR and MSSC problems, through which we prove NP-hardness of the generalized CLR problem. Column generation is an algorithmic framework for solving large-scale linear and integer programs. Vanderbeck and Wolsey \cite{Vanderbeck1996} and Barnhart \textit{et al.} \cite{Barnhart1998} overview column generation for solving large integer program. We design a column generation (CG) algorithm for the generalized CLR problem using its set partitioning formulation. The corresponding pricing problem is a mixed integer quadratic program, which we show to be NP-hard. To handle larger instances in the column generation framework, we also propose a heuristic algorithm, referred to as the CG Heuristic algorithm. This heuristic algorithm, inspired by Bertsimas and Shioda \cite{Bertsimas2007}, first clusters entities to a small number of groups and then performs our column generation algorithm on these groups of entities. In addition, we propose a metaheuristic algorithm, named the GA-Lloyd algorithm, which uses an adapted Lloyd's clustering algorithm to find locally optimal partitions and relies on the genetic algorithm (GA) to escape local optimums. Furthermore, we introduce a two-stage approach, used frequently in practice due to its simplicity, which performs clustering first and regression second.  We test our algorithms using real-world data from a large retail chain.  We compare the performance of the GA-Lloyd, the CG Heuristic, and the two-stage algorithms on two larger instances with 66 and 337 SKUs, corresponding to two representative subcategories under the retailer's product hierarchy. We observe that the GA-Lloyd algorithm performs much better than the two-stage algorithm. The CG Heuristic algorithm is able to produce slightly better results than the GA-Lloyd algorithm for smaller instances, but at the cost of much longer running time. The GA-Lloyd algorithm performs the best and identifies distinctive and meaningful seasonal patterns for the tested subcategories. In addition, we find that the column generation algorithm is able to solve the SKU clustering problem with at most 20 SKUs to optimality within reasonable computation time.  We benchmark the performance of the GA-Lloyd and CG Heuristic algorithms against the optimal solutions obtained by the column generation algorithm to find that both algorithms obtain close to optimal solutions.

The contributions of our work are as follows.
\begin{enumerate}[noitemsep]
\item We are the first to model and solve the SKU clustering problem, commonly encountered in retail predictive modeling, through generalized CLR.
\item We propose four heuristic algorithms for the generalized CLR problem, including the CG Heuristic algorithm, the GA-Lloyd algorithm, the two-stage approach, and a variant of Sp{\"a}th algorithm.
\item We propose an exact column generation algorithm that enables us to evaluate the performance of the heuristic algorithms.
\item We prove NP-hardness of the generalized CLR problem and NP-completeness of the pricing problem of the column generation algorithm.
\end{enumerate}

Note that the number of clusters is a parameter in the generalized CLR problem that needs to be decided by user beforehand or by enumeration. Although we provide comparison of models with different number of clusters for real-world data in Section \ref{subsec_compare_seasonality}, it is not straightforward to develop a universal rule for deciding the number of clusters. This is also a hard task for MSSC and CLR. The AIC or BIC criteria did not give a reasonable number of clusters for the data set we used for the experiment. They gave more than three times the number of hand-picked number of clusters that work in practice. Hence, in this paper, we assume that the target number of clusters is given in advance.

\begin{figure}[ht]
\center
  \includegraphics[scale=0.5]{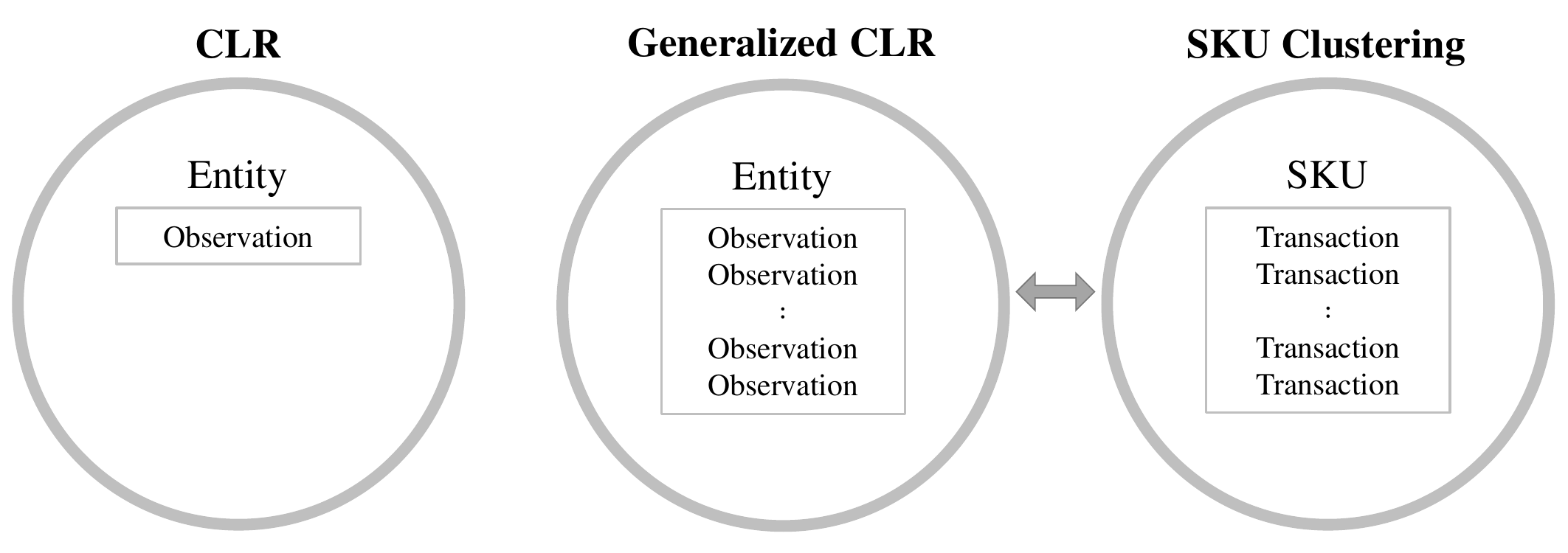}\\
  \caption{Comparison of Problems}
  \label{Fig:compare}
\end{figure}

In Figure \ref{Fig:compare}, we summarize and compare the terms in the CLR, generalized CLR and the SKU clustering problem. While CLR only has entities, generalized CLR allows multiple observations per entity. The CLR problem can be thought of as the generalized CLR with one observation per entity. Note that entity and observation in generalized CLR are the SKU and transactions in the SKU clustering problem, respectively.

The rest of the paper is organized as follows.  In Section \ref{Sec:Formulations}, we introduce both the mixed integer quadratic program and the set partitioning formulations of the generalized CLR problem. We draw the connection between the generalized CLR and MSSC problems, and prove NP-hardness of the former through this connection. In Section \ref{Sec:Algorithms}, we present the exact column generation algorithm, the CG Heuristic algorithm, the GA-Lloyd heuristic algorithm, the two-stage algorithm, and a variant of the Sp{\"a}th algorithm. The pricing problem of the column generation algorithm is shown to be NP-complete. In Section \ref{Sec:Experiments}, we present numerical experiments to test the performance of all proposed algorithms. 
%Edit by Conrado
% using real-world data from a large retail chain. We conclude the introduction with a literature review.
The literature review is discussed next.

%\section{Literature Review}\label{Sec:Literature}
\subsection{Literature Review}

To the best of authors' knowledge, no previous work has been conducted that comprehensibly tackles the generalized CLR problem. However, an extensive collection has been proposed for the typical CLR problem, which can potentially be adapted to tackle the generalized CLR problem. 

The algorithms proposed for the typical CLR problem are mainly heuristics bearing close similarity to the algorithms for the MSSC problem. For example, Sp{\"a}th \cite{Spath1979} proposes an exchange algorithm which, starting from some initial clusters, exchanges two items between two clusters if a cost reduction is observed in the objective function. DeSarbo \cite{DeSarbo1989} presents a simulated annealing method to escape local minimums. Muruzabal \textit{et al.}  \cite{Muruzabal2012} used a self organizing map to perform clusterwise regression.

On mathematical programming-based heuristics, Lau \textit{et al.} \cite{Lau1999} propose a nonlinear programming formulation that it is solved approximately using commercial solvers with no guarantee to find a global optimum. Their algorithm's performance depends heavily on the initial clusters. This initial-cluster dependency is overcome by the K-harmonic means clustering algorithm proposed by Zhang \cite{Zhang2003}. Moreover, Bertsimas and Shioda \cite{Bertsimas2007} introduce a compact mixed-integer linear formulation for a slight variation of the CLR problem with the sum of the absolute error as the objective. Their algorithm first divides entities into a small number of clusters, and then feeds these clusters into their mixed integer program. 

For exact approaches to CLR, Carbonneau \textit{et al.} \cite{Carbonneau2011} proposed a mixed logical-quadratic programming formulation by replacing big M constraints with the logical implication of the constraints. Carbonneau \textit{et al.} \cite{Carbonneau2012} proposed an iterative algorithm based on sequencing the data and repetitive use of a branch and bound algorithm. Carbonneau \textit{et al.} \cite{Carbonneau2014} proposed a column generation based algorithm based on \cite{Carbonneau2011} and \cite{Carbonneau2012}.

There are two key differences between these works and the one we propose in this paper. First, we provide both a quadratic mixed-integer program formulation and a set partition formulation of the generalized CLR problem. The former is a generalization of the formulation in \cite{Bertsimas2007}, and the latter is the set partitioning formulation for generalized CLR (recently, Carbonneau \textit{et al.} \cite{Carbonneau2014} have proposed a set partitioning formulation for CLR). Second, we propose two new heuristics, namely the CG Heuristic algorithm and the GA-Lloyd algorithm for the generalized CLR problem.

%Third, we have tested our algorithms on a real retail data set while the other papers do not handle practical data.

There is another stream of research for the CLR problem that assumes a distribution function for regression errors where each entity is assigned to each cluster with a certain probability, i.e., using ``soft'' assignments. For example, DeSarbo and Cron \cite{DeSarbo1988} propose a finite conditional mixture maximum likelihood methodology, which assumes normal distribution for regression errors and is solved through the expectation maximization algorithm. Since then, a large number of mixture regression models have been developed, including probit and logit mixture regression models as examples. Lau \textit{et al.} \cite{Lau1999} compare the performance of the expectation maximization algorithms with their nonlinear programming-based algorithm. Hennig \cite{Hennig2000} investigate idenfiability of model-based clusterwise linear regression for consistent estimate of parameters. D'Urso \textit{et al.} \cite{Durso2010} proposed to integrate fuzzy clustering and fuzzy regression. A recent work of Ingrassia \textit{et al}. \cite{Ingrassia2014} uses linear $t$ cluster-weighted models for clustering regression. These model-based approaches allow residual variances to differ between clusters, which the least squares approaches do not allow. In the soft assignment setting, an entity can be assigned to the cluster of highest probability. We restrict the scope of our review and comparison to least squares approaches because the objective functions are different. The reader is referred to Wedel and DeSarbo \cite{WedelDeSarbo1994} and Hennig \cite{Hennig1999} for reviews.

The algorithms for the MSSC problem are instructive to solving the CLR problem. There are abundant papers for solving the MSSC problem. Hansen and Jaumard \cite{Hansen1997} survey various forms of clustering problems and their solution methods, including MSSC, from a mathematical programming point of view. In their survey, solution methods for the MSSC problem include dynamic programming, branch-and-bound, cutting planes, and column generation methods. All these algorithms do not scale well to large size instances or in higher dimensional spaces. Heuristics are also considered, including Lloyd's like algorithms (e.g., K-Means and H-Means) and metaheuristics such as simulated annealing, tabu search, genetic algorithms and variable neighborhood search. With respect to mathematical programming approaches, du Merle \textit{et al.} \cite{DuMerle2000} propose an interior point algorithm to exactly solve the MSSC problem. Aloise \textit{et al.} \cite{Aloise2012} improve the algorithm of du Merle \textit{et al.}  \cite{DuMerle2000} by exploiting the geometric characteristics of clusters, which enables them to solve much larger instances.

\section{Problem Formulations}\label{Sec:Formulations}
\subsection{Mixed Integer Quadratic Program Formulation}
We first provide a mixed integer quadratic formulation for the generalized CLR problem. This formulation reveals a close connection between the generalized CLR and MSSC problems, which enables us to show that the generalized CLR problem is NP-hard.

Consider set $\{1,2,...,I\}$ of $I$ entities. Each entity $i\in I$ has $L$ observations of dependent variable $\textbf{y}_i = (y_{i1}, y_{i2}, ..., y_{iL})$, and $J$ independent variables $\textbf{x}_{i1}, \textbf{x}_{i2},..., \textbf{x}_{iJ}$ with $\textbf{x}_{ij}  = (x_{ij1}, x_{ij2}, ..., x_{ijL})$ for any $ j \in [J]$. In practice the number of entities $L$ depends on $i$, but we do not show this dependency for improved readability. (For each integer $g$ we introduce $[g] = \{1,...,g\}$.) Observation $y_{il}$ is associated with independent variables $x_{i1l}, x_{i2l}, ..., x_{iJl}$. Note that vectors are represented in bold symbols. We want to divide these $I$ entities into a partition $\boldsymbol{C}$ of $K$ clusters where $\boldsymbol{C} = (C_1, C_2, ..., C_K)$, $C_i \cap C_j = \emptyset$ for any $i \neq j$, and $\mathop{\cup}\limits_{k \in [K]}C_k = [I]$. The minimum size of a cluster is $n$, which is set by the user. This implies $|C_k| \geq n$ for any $k \in [K]$ where $|C_k|$ denotes the cardinality of cluster $C_k$. Note that the number of observations pertaining to a cluster is at least $nL$. The minimum size constraints are imposed to ensure that there are enough observations for each cluster. Further, in order to avoid regression models with zero error, we require $L \cdot n > J +1$. We also require $I \geq K \cdot n$ such that there is always a feasible solution. The generalized CLR problem is formulated as follows:
\begin{alignat}{2}
\min \sum_{i=1}^I\sum_{l=1}^{L}t_{il}^2& \label{Eq:MixedIntegerQuadratic}\\
t_{il} - (y_{il} - \sum_{j=1}^J\beta_{kj}x_{ijl}) + M(1 - z_{ik}) & \geq  0 \quad & &i \in [I]\text{, } k \in [K]\text{, } l \in [L]\label{constraint:quadratic:regression1} \\
t_{il} + (y_{il} - \sum_{j=1}^J\beta_{kj}x_{ijl}) + M(1 - z_{ik}) & \geq  0 \quad & &i \in [I]\text{, } k \in [K]\text{, } l \in [L]\label{constraint:quadratic:regression2} \\
\sum_{k=1}^K z_{ik} & =1 \quad & &i \in [I] \label{constraint:quadratic:assignment} \\
\sum_{i=1}^I z_{ik} & \geq n \quad & &k \in [K] \label{constraint:quadratic:minimumSize} \\
z_{ik} &\in \{0,1\} \quad & &i \in [I]\text{, } k \in [K] \nonumber \\
t_{il} &\geq 0 \quad & &i \in [I]\text{, } l \in [L] \nonumber\\
\beta_{kj}&\text{\ unconstrained} \quad & &k \in [K]\text{, } j \in [J], \nonumber
\end{alignat}
where $z_{ik}$ is a binary variable, which is equal to one if and only if entity $i$ is assigned to cluster $C_k$. Value $M$, referred to as big $M$ in the optimization literature, is a large positive constant. Due to constraints \eqref{constraint:quadratic:regression1} and \eqref{constraint:quadratic:regression2}, $t_{il}$ is equal to the absolute error for the corresponding observation $y_{il}$ in the optimal solution, and $\boldsymbol{\beta}_k = (\beta_{k1}, \beta_{k2},...,\beta_{kJ})$  are the regression coefficients for cluster $C_k$, which are decision variables. The role of $M$ is to enforce constraints \eqref{constraint:quadratic:regression1} and \eqref{constraint:quadratic:regression2} only when they are needed (entity $i$ is assigned to cluster $k$). In detail, if $z_{ik} = 1$, then we have $t_{il} -  (y_{il} - \sum_{j=1}^J\beta_{kj}x_{ijl}) \geq 0$, and $t_{il} + (y_{il} - \sum_{j=1}^J\beta_{kj}x_{ijl})\geq 0$, which implies $t_{il} =  |(y_{il} - \sum_{j=1}^J\beta_{kj}x_{ijl})|$ because we are minimizing  the sum of $t^2_{il}$. If $z_{ik} = 0$, constraints \eqref{constraint:quadratic:regression1} and \eqref{constraint:quadratic:regression2} require $t_{il}$ to be greater than a negative number, which holds trivially due to the existence of the nonnegativity constraint on $t_{il}$. Constraint \eqref{constraint:quadratic:assignment} requires that every entity is assigned to one cluster, and \eqref{constraint:quadratic:minimumSize} imposes the limit on the cardinality of each cluster.

Unlike the CLR problem, the generalized CLR allows each entity to have more than one observation, which implies that $L$ can be greater than one. The mixed integer linear program formulation for the CLR problem in Bertsimas and Shioda \cite{Bertsimas2007} has $L$ equal to one, and does not have the cluster cardinality constraint \eqref{constraint:quadratic:minimumSize}. Besides, their objective function is the sum of the absolute errors while ours is the sum of squared errors.

Our SKU clustering problem based on the seasonal effects can be modeled as the generalized CLR problem. The entities to cluster are SKUs. The response variable $\textbf{y}_i$ corresponds to a vector of weekly sales for SKU $i$. The independent variables $\textbf{x}_{i}$'s include promotional predictors such as promotion mechanisms,  percentage discount, and seasonal dummies for SKU $i$.

Aloise \textit{et al.} \cite{Aloise2009} showed NP-hardness of the MSSC problem in a general dimension when the number of clusters is two. General dimension means that the size of the vectors to be clustered is not a constant but part of the input data. A similar statement can be made for the generalized CLR problem with the proof available in Appendix \ref{appendix_proof_theorems}.
\begin{theorem}\label{Theorem:CLR}
The generalized CLR problem with two clusters in a general dimension is NP-hard.
\end{theorem}

With the formulation presented by \eqref{Eq:MixedIntegerQuadratic}--\eqref{constraint:quadratic:minimumSize}, we can solve the generalized CLR problem using any commercial optimization software that can handle quadratic mixed integer programs. However, this formulation suffers from two drawbacks, which makes it intractable for large instances. The first one relates to big $M$. Optimality of the solution and efficiency of integer programming solvers depend on a tight value of $M$. Unlike multiple linear regression, where obtaining a valid value of $M$ is possible \cite{Park13}, it is not trivial to calculate a valid value of $M$ in \eqref{constraint:quadratic:regression1} and \eqref{constraint:quadratic:regression2} for the generalized CLR or CLR. When $z_{ik} = 0$, $\beta_{kj}$'s are not from the cluster that entity $i$ belongs to, and the residual $t_{il}$ can be arbitrarily large. Carbonneau \textit{et al.} \cite{Carbonneau2011} provide an empirical result that a big M based MIP formulation for CLR sometimes fails to guarantee optimality of CLR for the data sets they consider. The second one involves the symmetry of feasible solutions. Any permutation of clusters yields the same solution, yet it corresponds to different decision variables. Symmetry unnecessarily increases the search space, and renders the solution process inefficient. To overcome the symmetry problem, we propose a set partitioning formulation, which has already been used for the CLR problem in \cite{Carbonneau2014}.

\subsection{Set Partitioning Formulation}\label{subsec:SetPartition}

Let $\mathscr{S}$ denote the set of all clusters of entities with the cardinality equal to or greater than $n$, i.e., $\mathscr{S} = \{S \subseteq [I], |S| \geq n\}$. Let $a_{iS}$ equal to one if entity $i$ belongs to cluster $S$, and equal to zero otherwise. Let $c_S$ denote the cost of cluster $S$, which is equal to the sum of squared errors when performing the regression over cluster $S$. Introducing binary variables
\begin{equation*}
z_S=\left\{
\begin{array}{rl}
1&\text{ if cluster }S\text{ is selected,} \\
0&\text{\ otherwise,}
\end{array} \right.
\end{equation*}
the generalized CLR problem can be formulated as:
\begin{alignat}{2}
\min \sum_{S\in \mathscr{S}} &c_Sz_S \label{Eq:MasterProblem}\\
\sum_{S\in \mathscr{S}} z_S &= K \label{constraints:partition:group}\\
\sum_{S\in \mathscr{S}} a_{iS}z_S &= 1 \quad & &i \in [I] \label{constraints:partition:assignment}\\
z_S &\in \{0,1\}\quad & &S \in \mathscr{S}.\nonumber
\end{alignat}
Constraint \eqref{constraints:partition:group} ensures that the number of clusters in the partition is $K$ and constraint \eqref{constraints:partition:assignment} guarantees that each entity occurs in only one cluster within the partition.

\section{Algorithms}\label{Sec:Algorithms}
\subsection{Column Generation (CG) Algorithm}
%Edit by Conrado
The set partitioning formulation has an exponential number of binary variables. It is very challenging to solve even its linear programming relaxation because there are so many decision variables. To solve large-scale linear and integer programs, column generation algorithms have been used in the literature. The reader is referred to Vanderbeck and Wolsey \cite{Vanderbeck1996} and Barnhart \textit{et al.} \cite{Barnhart1998} for reviews of column generation for solving large-scale integer programs. In our work, we employ column generation to handle its linear programming relaxation. At the high level, column generation can be understood as iteratively expanding set $\bar{\mathscr{S}}$ (a subset of $\mathscr{S}$) in \eqref{Eq:MasterProblem} - \eqref{constraints:partition:assignment} by adding attractive candidate cluster $S$ to $\bar{\mathscr{S}}$. The key challenge is how to select $S$. The word column is used because adding cluster $S$ to $\bar{\mathscr{S}}$ is equivalent to adding a column in the matrix form of \eqref{Eq:MasterProblem} - \eqref{constraints:partition:assignment}.

The column generation algorithm, referred to as the CG algorithm, starts by solving the restricted master problem which has the same formulation as the master problem \eqref{Eq:MasterProblem}-\eqref{constraints:partition:assignment}, but with set $\mathscr{S}$ replaced by $\bar{\mathscr{S}}$, a smaller subset of columns. Recall that a column represents a cluster (subset of entities $[I]$). We start the algorithm with small candidate clusters rather than $\mathscr{S}$, the set of all possible subsets of $[I]$. The algorithmic framework is presented in Algorithm \ref{alg:exactSCG1}, which follows the general column generation scheme. In Line 1, the initial subset of columns in $\bar{\mathscr{S}}$ are randomly generated. In detail, we start from $K$ empty clusters. Then, we randomly assign each entity to one of the $K$ clusters using a uniform random number. Hence, after Line 1, we have $K$ clusters and $|\bar{\mathscr{S}}| = K$ for the generation procedure. In Line 3, optimal dual variables are obtained by solving the restricted master problem and then serve as input to the pricing problem, which will be introduced hereafter, to calculate the smallest reduced cost column. In Line 4, the pricing problem returns a column with the smallest reduced cost. In Lines 5-10, if the reduced cost is nonnegative, then we conclude that the master problem is solved optimally. Otherwise, we add the column with the smallest reduced cost to the restricted master problem and repeat the process.

\begin{algorithm}[ht] 
\caption{CG} \label{alg:exactSCG1}
\begin{algorithmic}[1]
%\State Initialize $\bar{\mathscr{S}}^{(0)} \subset \mathscr{S}$, $\boldsymbol{\delta}^{(0)}$, $\boldsymbol{\xi}^{(0)}$
%$\boldsymbol{\xi}^{(0)}$:  \\
\STATE Randomly generate $\bar{\mathscr{S}}$ (a small subset of $\mathscr{S}$)
\WHILE{ not optimal }
\STATE Solve master problem \eqref{Eq:MasterProblem} -- \eqref{constraints:partition:assignment} and obtain dual solution
\STATE Get a new cluster by solving pricing problem with input of dual solution from Line 3
\IF {the reduced cost is nonnegative}
    \STATE The algorithm is complete with the optimal partition of clusters
\ELSE
	\STATE Add the cluster from Line 4 to the master problem
\ENDIF
\ENDWHILE
\end{algorithmic}
\end{algorithm}

%\begin{align}
%&\min \sum_{S\in \mathscr{S}} c_Sz_S \label{Eq:LROriginal}\\
%\text{s.t.\ \ }&\sum_{S\in \mathscr{S}} z_S \leq K \nonumber\\
%&\sum_{S\in \mathscr{S}} a_{iS}z_S \geq 1 \text{\ \ }\forall i = 1,...,I \nonumber\\
%&z_S \geq 0\text{\ \ } \forall S \in \mathscr{S} \nonumber.
%\end{align}
%Note that we do not need the constraints $z_S \leq 1$ because it is never optimal to have $z_S$ greater than one.

\subsubsection*{The pricing problem}

The pricing problem can be stated as follows. Let $\upsilon$ be the dual variable for constraint \eqref{constraints:partition:group}, and $\pi_i$'s be the dual variables for constraint \eqref{constraints:partition:assignment}. The reduced cost for cluster $S$ is $d_S=c_S-\upsilon-\sum_i\pi_i a_{iS}$, and thus the pricing problem reads:
\begin{align}
\min_{|S|\geq n, \boldsymbol{\beta}} \sum_{i \in S}\sum_{l=1}^L(y_{il} - \sum_{j=1}^J x_{ijl}\beta_j)^2 - \sum_{i \in S} \pi_i.\label{Eq:PricingNPFormulation}
\end{align}
Note that we omit the subtraction of $\upsilon$ in the formulation because it is a constant which does not change the optimal solution.

\begin{theorem}\label{TH:PricingNPComplete}
The pricing problem as stated in \eqref{Eq:PricingNPFormulation} is NP-complete.
\end{theorem}

The proof is available in Appendix \ref{appendix_proof_theorems}. Introducing binary variables
\begin{equation*}
z_i=\left\{
\begin{array}{rl}
1&\text{ if } i \in S\text{,} \\
0&\text{\ otherwise,}
\end{array} \right.
\end{equation*}
the pricing problem can be formulated as a mixed integer quadratic program:
\begin{alignat}{2}
\min \sum_{i = 1}^I\sum_{l = 1}^{L}t_{il}^2-\sum_{i = 1}^I{\pi_i
z_{i}}& \label{Eq:Pricing}\\
t_{il}-(y_{il}-\sum_{j=1}^J\beta_j x_{ijl})+M(1-z_i)&\geq 0  \text{\ \ }& &i \in [I] \text{ , } l \in [L] \label{constraints:pricing:Residual1}\\
t_{il}+(y_{il}-\sum_{j=1}^J\beta_j x_{ijl})+M(1-z_i)&\geq 0  \text{\ \ }& &i \in [I] \text{ , } l \in [L] \label{constraints:pricing:Residual2}\\
\sum_{i=1}^I z_i &\geq n \label{constraint:pricing:MinSize} \\
t_{il}&\geq 0 \text{\ \ } & &i \in [I] \text{ , } l \in [L]\nonumber\\
z_i&\in \{0,1\}\text{\ \ } & &i \in [I],\nonumber
\end{alignat}
where $M$ is a large positive constant and is assumed to be valid (does not cut an optimal solution). We can use the same approach from Section 2.1 to set up a valid value for $M$. A feasible solution's SSE can be a valid value. By using similar arguments as those for constraints \eqref{constraint:quadratic:regression1} and \eqref{constraint:quadratic:regression2}, $t_{il}$ is the absolute error for the corresponding observation $y_{il}$ in the optimal solution if $i \in S$, and it is zero otherwise. The difference from the pricing problem in \cite{Carbonneau2014} is that \eqref{Eq:Pricing}--\eqref{constraint:pricing:MinSize} is based on big M constraints and is for the generalized CLR, while Carbonneau \textit{et al.} \cite{Carbonneau2014} used logical implications of the constraints for the CLR problem.

In the column generation algorithm, \eqref{Eq:Pricing}--\eqref{constraint:pricing:MinSize} are solved. Recall that reduced cost for cluster $S$ is $d_S=c_S-\upsilon-\sum_i\pi_i a_{iS}$. It is easy to see that value $d_S + \upsilon$ is equivalent to to the value of \eqref{Eq:Pricing} with $z_i = 1$ for $i \in S$ and 0 otherwise. This follows from the fact that $d_S + \upsilon = c_S - \sum_{i\in S} pi_i$ and $c_S$ is modeled by variables $t$.

\noindent \textbf{Example} Let us consider a data set with 4 entities and suppose $n = 1$. Then, we have \begin{center}
$\mathscr{S} = \big\{ \{1\}, \{2\}, \cdots, \{1,2\}, \{1,3\}, \cdots, \{1,2,3\}, \{1,2,4\}, \cdots, \{1,2,3,4\} \big\}$,
\end{center}
where $|\mathscr{S}| = 15$. In Algorithm \ref{alg:exactSCG1}, suppose we start with subset $\bar{\mathscr{S}} = \big\{\{1\}, \{2\}, \{3\}, \{2,3\}, \{3,4\}, \{1,2,4\} \big\}$ of $\mathscr{S}$, which is a set of initial candidate clusters. The master problem in Line 3 picks the best combination of the candidate clusters that has minimum total SSE. Suppose we obtain $\{1\}, \{2\}, \{3,4\}$ in Line 3 together with the associated dual solution. Here we assume that the solution is integral, albeit this might not always be the case. In Line 4, the pricing problem is solved to search if there exists a candidate cluster not in $\bar{\mathscr{S}}$ that can improve the current best solution $\{1\}, \{2\}, \{3,4\}$. Suppose the pricing problem returns $\{1,4\}$ with a negative reduced cost. In Line 8, $\bar{\mathscr{S}}$ is updated to  $\bar{\mathscr{S}} = \big\{\{1\}, \{2\}, \{3\}, \{1,4\}, \{2,3\}, \{3,4\}, \{1,2,4\} \big\}$.

%%%%=================================================================
%%%%=================================================================
\subsubsection*{Column generation stabilization schemes}
%%%% Instead of directly solving (\ref{Eq:LROriginal}), we perturb it by adding bounded surplus and slack variables, and penalize violations of its constraints in the objective function, which leads to the following formulation:
%\begin{align}
%&\min \sum_{S\in \bar{\mathscr{S}}} c_Sz_S - \delta_i^{-}q_i^{-} + \delta_i^{+}q_i^{+}\label{Eq:SCG}\\
%\text{s.t.\ \ }&\sum_{S\in \bar{\mathscr{S}}} z_S \leq K \nonumber\\
%&\sum_{S\in \bar{\mathscr{S}}} a_{iS}z_S - q_i^{-} + q_i^{+}\geq 1 \text{\ \ }\forall i = 1,...,I \nonumber\\
%&z_S \geq 0\text{\ \ } \forall S \in \bar{\mathscr{S}}\\ \nonumber
%&0\leq q_i^{-}\leq \xi^{-} \text{\ \ }\forall i = 1,...,I \nonumber\\
%&0\leq q_i^{+}\leq \xi^{+} \text{\ \ }\forall i = 1,...,I \nonumber
%\end{align}
%
If the optimal solution obtained by CG is not integral, branching would have to be performed, i.e., a fractional variable $z_S$ needs to be selected and two new problems created, the first one would impose $z_S = 0$ and the other one $z_S = 1$. However, the extensive evaluation conducted on Algorithm \ref{alg:exactSCG} revealed that no fractional solutions were provided by Algorithm \ref{alg:exactSCG}. For this reason in the remainder we focus on column generation for solving the LP relaxation and not branching. Column generation is known to exhibit the tailing-off effect and for this reason we employ stabilized column generation of du Merle \textit{et al.} \cite{duMerle1999}.

The stabilized column generation algorithm for solving the CLR problem is illustrated in Algorithm \ref{alg:exactSCG}. The algorithm takes input of stabilization parameters $\boldsymbol{\delta}^{(0)}$ and $\boldsymbol{\xi}^{(0)}$, and maximum allowed iterations $k^{max}$. In Line 1, we start with a set $\bar{\mathscr{S}}^{(0)}$ of  initial clusters of entities. The generation procedure is identical to the one in Line 1 of Algorithm 1. For iteration $k$, in Line 3, we solve the stabilized master problem and get the optimal solution $(\boldsymbol{z}^{(k)}, \boldsymbol{q}^-,\boldsymbol{q}^+)$ and its corresponding dual solution $(\boldsymbol{\pi}^{(k)}, \upsilon)$, which provides input parameters for the pricing problem. The stabilized master problem additionally includes parameters $\boldsymbol{\delta}^{(k)}$, $\boldsymbol{\xi}^{(k)}$ and variables $\boldsymbol{q}^-$, $\boldsymbol{q}^+$ but is very similar to \eqref{Eq:MasterProblem} - \eqref{constraints:partition:assignment}. See Appendix \ref{appendix_Restricted_Master_Problem} for the actual formulation. By solving the pricing problem, we get a new cluster $S^{(k)}$ in Line 4. The reduced cost corresponding to this new cluster is equal to $c_S - \boldsymbol{a}_S^\intercal\boldsymbol{\pi}^{(k)} - \upsilon$, where $c_S$ is the sum of squared residuals when performing regression over this cluster, and $a_{iS} = 1$ if and only if $i \in S$. In Lines 5-6, if the reduced cost is nonnegative and $\boldsymbol{q}^-$ and $\boldsymbol{q}^+$ are equal to zero, then the algorithm is complete with the optimal partition of clusters defined by $\boldsymbol{z}^{(k)}$. Otherwise, in Lines 8-12, we update  $\bar{\mathscr{S}}$ and then if the reduced cost $c_S - \boldsymbol{a}_S^\intercal\boldsymbol{\pi}^{(k)} - \upsilon$ is nonnegative, we update the stabilization parameters $\boldsymbol{\delta}^{(k)}$, and $\boldsymbol{\xi}^{(k)}$.

%\begin{algorithm*}[t] 
\begin{algorithm}[ht] 
\caption{CG($\boldsymbol{\delta}^{(0)}$, $\boldsymbol{\xi}^{(0)}, k^{max}$)} \label{alg:exactSCG}
\begin{algorithmic}[1]
%\State Initialize $\bar{\mathscr{S}}^{(0)} \subset \mathscr{S}$, $\boldsymbol{\delta}^{(0)}$, $\boldsymbol{\xi}^{(0)}$
\STATE $k\gets 0$, randomly generate $\bar{\mathscr{S}}^{(0)}$ (a small subset of $\mathscr{S}$)
\WHILE{$k < k^{max}$}
%\REPEAT
\STATE $(\boldsymbol{z}^{(k)}, \boldsymbol{q}^-,\boldsymbol{q}^+; \boldsymbol{\pi}^{(k)}, \upsilon) \gets$ solve the stabilized master problem \eqref{Eq:RestrictedMasterProblem} - \eqref{restr222} given in Appendix \ref{appendix_Restricted_Master_Problem}
\STATE $ S^{(k)} \gets$ solve the pricing problem with $\boldsymbol{\pi}^{(k)}$
\IF {$\boldsymbol{a}_S^\intercal\boldsymbol{\pi}^{(k)} + \upsilon\leq c_S$ and $\boldsymbol{q}^- = \boldsymbol{q}^+ = \boldsymbol{0}$}
    \STATE $\boldsymbol{z}^* \gets \boldsymbol{z}^{(k)}$, and \textbf{stop} %$\text{End} \gets \text{True}$
\ELSE
    \STATE $\bar{\mathscr{S}}^{(k+1)} \gets \bar{\mathscr{S}}^{(k)}\cup S^{(k)}$
    \IF {$\boldsymbol{a}_S^\intercal\boldsymbol{\pi}^{(k)} + \upsilon\leq c_S$}
        \STATE $\boldsymbol{\delta}^{(k+1)} \gets \text{Update }(\boldsymbol{\delta}^{(k)})$
        \STATE $\boldsymbol{\xi}^{(k+1)} \gets \text{Update }(\boldsymbol{\xi}^{(k)})$
    \ENDIF
\ENDIF
  \STATE $k \gets k+1$
%\UNTIL{$k > MaxIter$}
\ENDWHILE
\end{algorithmic}
\end{algorithm}

%%%%%%%%%%%%%%%%%%%%%%%%%%%%%%%%%%%%%%%%%%%%%%%%%%%%%%%%%%%%%%%%%%%%%%%%%%%%%%%%%%%%%%%%%%%%%%%%%%%%%%%%%%%%%%%%%%%%%%%%%%%%%%
%%%%%%%%%%%%%%%%%%%%%%%%%%%%%%%%%%%%%%%%%%%%%%%%%%%%%%%%%%%%%%%%%%%%%%%%%%%%%%%%%%%%%%%%%%%%%%%%%%%%%%%%%%%%%%%%%%%%%%%%%%%%%%
%%%%%%%%%%%%%%%%%%%%%%%%%%%%%%%%%%%%%%%%%%%%%%%%%%%%%%%%%%%%%%%%%%%%%%%%%%%%%%%%%%%%%%%%%%%%%%%%%%%%%%%%%%%%%%%%%%%%%%%%%%%%%%

\subsection{CG Heuristic Algorithm}

The numerical experiments introduced later reveal that the column generation algorithm does not scale well to problems with a large number of entities. 
%To potentially circumvent the scalability problem, we propose the following heuristic method, called the CG heuristic algorithm, 
To overcome the scalability problem, we propose a heuristic method called the CG Heuristic algorithm that relies on column generation.

The CG Heuristic algorithm first finds a partition with a large number of clusters by neglecting the cardinality constraint. In the second step, we combine the clusters by considering unions to obtain exactly $K$ clusters while obeying the cardinality constraint, which is a slight variant of column generation. We refer to the intermediate clusters from the first part, which are the input to the column generation algorithm in the second part, as groups. 

The algorithmic framework is presented in Algorithm \ref{alg:CGHeur}. We require that $R>K$ since the second step is to combine $R$ groups into $K$ clusters. Lines 1-5 represent the first step to create $R$ groups and Line 6 represents the second step to find a solution to the original problem. Line 1 follows the same procedure as Line 1 of Algorithm 1, except that we have R groups instead of K. Lines 2-5 are basic and do not need further explanations. It yields $R$ ``low cost'' groups. Since $R > K$, in Line 6 we combine some groups so that we end up with exactly $K$ clusters, each one with cardinality at least $n$. This regrouping of groups is performed in an optimal way by using the column generation framework.

\begin{algorithm}[ht] 
\caption{CG Heuristic ($R$)} \label{alg:CGHeur}
\begin{algorithmic}[1]
\REQUIRE $R (>K)$
%\State Initialize $\bar{\mathscr{S}}^{(0)} \subset \mathscr{S}$, $\boldsymbol{\delta}^{(0)}$, $\boldsymbol{\xi}^{(0)}$
\STATE Randomly generate $R$ groups
\WHILE{there is an update in groups}
\STATE Perform regression over each group $r \in [R]$ to obtain regression coefficients $\boldsymbol{\beta}_r$
\STATE For $i \in [I]$, reassign entity $i$ to group $r^*$, where $r^* = \argmin_r \sum_{l=1}^L(y_{il} - \sum_{j=1}^J \beta_{rj}x_{ilj})^2$
\ENDWHILE
\STATE Execute CG by treating each group as entity
\end{algorithmic}
\end{algorithm}

For Line 6, we need to revise the master and pricing problems in the following way when we cluster a group of entities instead of single entities. Suppose at the end of Line 5 we clustered $I$ entities into $R$ groups $\{G_1, G_2,...,G_R\}$, and then apply the column generation algorithm to the $R$ groups of entities. Let $\mathscr{S}_R$ be the set of all subsets $S$ of $[R]$ such that $|\mathop{\cup}\limits_{r \in S} G_r| \geq n$, and let $a_{rS} = 1$ if $r \in S$, and $a_{rS} = 0$ otherwise. To obtain the new master problem, we need to replace $\mathscr{S}$ with $\mathscr{S}_R$ and $a_{iS}$ with $a_{rS}$ in the master problem \eqref{Eq:MasterProblem}--\eqref{constraints:partition:assignment}. In addition, the range of constraints \eqref{constraints:partition:assignment} changes to $r \in [R]$.

%The master problem becomes the following:
%\begin{align*}
%&\min \sum_{S\in \mathscr{S}_R} c_Sz_S\\
%\text{s.t.\ \ }&\sum_{S\in \mathscr{S}_R} z_S=K\\
%&\sum_{S\in \mathscr{S}_R} a_{rS}z_S=1 \text{\ \ }\forall r = 1,...,R\\
%&z_S\in \{0,1\}\text{\ \ } \forall S \in \mathscr{S}_R.
%\end{align*}

We denote the dual variables of constraints \eqref{constraints:partition:assignment} in the master problem by $\pi_r$, and introduce the binary decision variables $z_r$ for $r \in [R]$ to indicate whether group $G_r$ is selected in the cluster with the minimum reduced cost. To obtain the new pricing problem, we need to replace $z_i$'s with $z_r$'s in the pricing problem \eqref{Eq:Pricing}--\eqref{constraint:pricing:MinSize}. Constraint \eqref{constraint:pricing:MinSize} is changed to $\sum_{r=1}^R |G_r|z_r \geq n$, and the range in constraints \eqref{constraints:pricing:Residual1} and \eqref{constraints:pricing:Residual2} now becomes $r \in [R] \text{ , } i \in G_r  \text{ , } l \in [L]$. The new pricing problem has the same number of constraints as the pricing problem \eqref{Eq:Pricing}--\eqref{constraint:pricing:MinSize}, however, it has only $R$ binary variables, comparing to $I$ such variables in the pricing problem ref{Eq:Pricing}--\eqref{constraint:pricing:MinSize}.

%%%%%%%%%%%%%%%%%%%%%%%%%%%%%%%%%%%%%%%%%%%%%%%%%%%%%%%%%%%%%%%%%%%%%%%%%%%%%%%%%%%%%%%%%%%%%%%%%%%%%%%%%%%%%%%%%%%%%%%%%%%%%%
%%%%%%%%%%%%%%%%%%%%Revise Column Generation by Generating Small Groups First%%%%%%%%%%%%%%%%%%%%%%%%%%%%%%%%%%%%%%%%%%%%%%%%%
%%%%%%%%%%%%%%%%%%%%%%%%%%%%%%%%%%%%%%%%%%%%%%%%%%%%%%%%%%%%%%%%%%%%%%%%%%%%%%%%%%%%%%%%%%%%%%%%%%%%%%%%%%%%%%%%%%%%%%%%%%%%%%

\subsection{GA-Lloyd Heuristic Algorithm}

Scientific works as those presented by Maulik and Bandyopadhyay \cite{Maulik2000} and Chang \textit{et al.} \cite{Chang2009}, effectively suggest to embed the concept of the Lloyd's algorithm, into a genetic search metaheuristic framework to find proper clusters for the MSSC clustering problem. Here we discuss our proposed adaptation of the GA-based Lloyd's clustering algorithms for solving the generalized CLR problem. 

For the Lloyd's algorithm part, a vector of regression coefficients $\boldsymbol{\beta}_k$ is used to represent cluster $k$, and an entity is recursively assigned to the cluster that gives the smallest sum of squared errors for this entity. The GA part helps escape local optimal solutions. The overall algorithmic framework is outlined in Algorithm \ref{alg:GALloyd}.

\begin{algorithm}[ht] 
\caption{GA-Lloyd} \label{alg:GALloyd}
\begin{algorithmic}[1]
\REQUIRE K, \textit{maxIter}, H, $p_m$

\STATE For each $h$ in $[H]$, create $\boldsymbol{C}(h)$ by randomly generating $K$ clusters
\WHILE{objective function value improved in the previous \textit{maxIter} iterations}
%\REPEAT
\STATE Randomly select parent chromosomes $h_1$ and $h_2$ using roulette wheel selection
\STATE Create child chromosomes $h_a$ and $h_b$ by performing crossover on $h_1$ and $h_2$
\STATE Mutation on $h \in \{h_a, h_b\}$
\STATE Obtain $\boldsymbol{C}(h_a)$  and $\boldsymbol{C}(h_b)$ based on Lloyd's algorithm, calculate $\gamma(h_a)$ and $\gamma(h_b)$
\IF {$\max\{\gamma(h_a), \gamma(h_b)\} < \min_{h \in [H]} \gamma(h)$}
    \STATE Replace $h^*$, $h^* = argmin_{h \in [H]} \gamma(h)$, with the chromosome with larger fitness among $h_a$ and $h_b$
\ENDIF
\ENDWHILE
\end{algorithmic}
\end{algorithm}

\begin{figure}[ht]
\center
  \includegraphics[scale = 0.6]{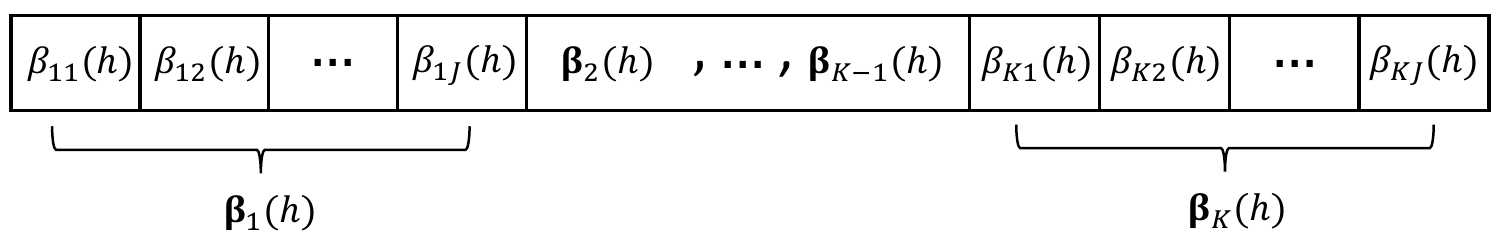}\\
  \caption{Encoding of Chromosome $h$}
  \label{Figure:Encoding}
\end{figure}

In Line 1, we start by randomly generating $H$ partitions $\boldsymbol{C}(1), ..., \boldsymbol{C}(H)$, each of which corresponds to $K$ clusters of entities with $\boldsymbol{C}(h) = \{C_1(h),C_2(h),...,C_K(h)\}$ for $h \in [H]$. The generation procedure is based on a uniform random number and is the same as the one in Line 1 of Algorithm 1. Any randomly generated partition $\boldsymbol{C}(h)$ has to satisfy the constraint that $|C_k(h)| \geq n$ for $k \in [K]$. A population $P$ consists of $H$ chromosomes, and chromosome $h$ is encoded as a vector $\boldsymbol{\beta}(h)$ of size $J\cdot K$. In any chromosome, the first $J$ genes represent the regression coefficients $\boldsymbol{\beta}_1(h)$ for the first cluster, and the next $J$ genes represent the regression coefficients $\boldsymbol{\beta}_2(h)$ for the second cluster, and so on. The encoding of chromosome $h$ is illustrated in Figure \ref{Figure:Encoding}. 

The regression coefficient $\boldsymbol{\beta}_k(h)$ is obtained by running regression over cluster $C_k(h)$. The fitness $\boldsymbol{\gamma}(h)$ of the chromosome $h$ is defined to be
\begin{equation}
\boldsymbol{\gamma}(h) =  \dfrac{1}{\sum_{k=1}^K\sum_{i\in C_k(h)} \sum_{l=1}^L(y_{il} - \sum_{j=1}^J \beta_{kj}(h)x_{ilj})^2}.
\label{Eq:Fit}
\end{equation}

\begin{figure}[ht]
\center
  \includegraphics[scale = 0.6]{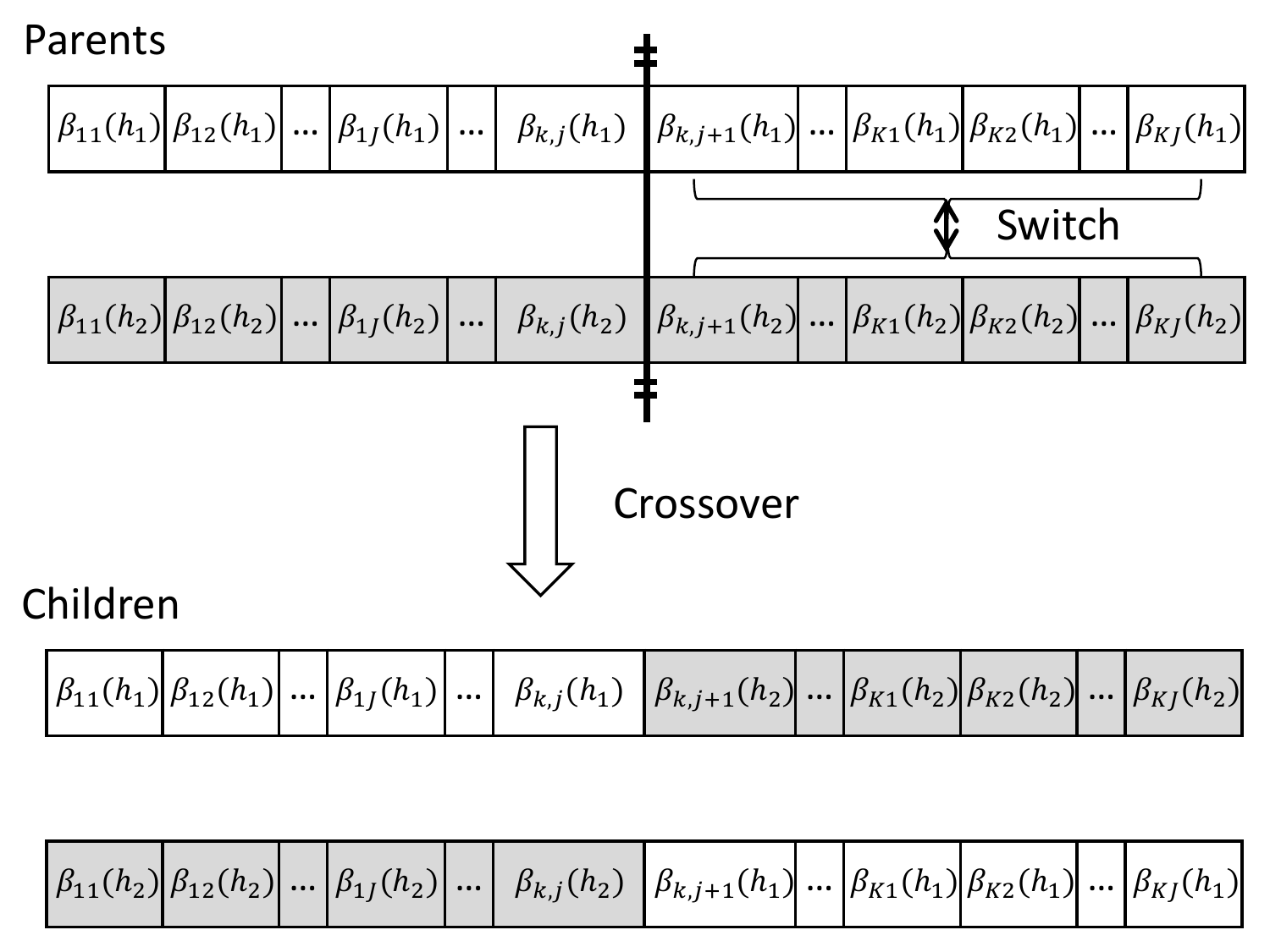}\\
  \caption{Crossover of Parent Chromosomes $h_1$ and $h_2$}
  \label{Figure:Crossover}
\end{figure}

We continue by performing the following genetic operations on the population of chromosomes iteratively until the number of iterations without improvement reaches a specified maximum number \textit{maxIter}. First, in Line 3, we randomly select two parent chromosomes $h_1$ and $h_2$ from population $P$ using roulette wheel selection. Chromosome $h$ is chosen with probability $\boldsymbol{\gamma}(h)/\sum_{g=1}^H \boldsymbol{\gamma}(g)$. Second, in Line 4, we perform crossover on chromosomes $h_1$ and $h_2$. We select a gene position as a random integer in the range of $[1,K\cdot J-1]$. We require this random integer to be no more than $K\cdot J-1$ so that there is at least one gene positioned to the right of it.  The portions of the chromosome lying to the right of this gene position are exchanged to produce two child chromosomes $h_a$ and $h_b$ encoded by $\boldsymbol{\beta}(h_a)$ and $\boldsymbol{\beta}(h_b)$. The crossover operation is illustrated in Figure \ref{Figure:Crossover}. 

Third, in Line 5, we perform mutation on these two child chromosomes. The mutation is performed on a child chromosome with a fixed probability $p$, where $p$ is a parameter.  A gene position with value $\upsilon$ is randomly picked from the child chromosome using a uniform random number. After mutation, it is changed to $\upsilon \pm 2 \delta v$ with equal probability if $\upsilon$ is not zero. Here $\delta$ is a random number with uniform distribution between zero and one. Otherwise, when $\upsilon$ is zero, it is changed to $\upsilon \pm 2 \delta$ with equal probability. In this way, the regression coefficients can take any real values after sufficient number of iterations. Next, in Line 6, we need to decode these two mutated child chromosomes to get the partitions $\boldsymbol{C}(h_a)$ and $\boldsymbol{C}(h_b)$ of clusters they represent. To decode the child chromosome $h_a$, we assign entity $i$ to cluster $C_{k^*_i}(h_a)$ for
\begin{equation*}
k_i^* = \argmin_k \sum_{l=1}^L(y_{il} - \sum_{j=1}^J \beta_{kj}(h_a)x_{ilj})^2
\end{equation*}
Then, we perform regression over each cluster of $\boldsymbol{C}(h_a)$ and $\boldsymbol{C}(h_b)$, and update the encoding of these two child chromosomes $\boldsymbol{\beta}(h_a)$ and $\boldsymbol{\beta}(h_b)$ with the resultant regression coefficients. Fitness $\gamma(H+1)$ and $\gamma(H+2)$ are calculated for the child chromosomes using \eqref{Eq:Fit}. In Lines 6-7, we replace the chromosome in population $P$ with the smallest fitness with the child chromosome with the smaller fitness if  $\max(\gamma(h_a),\gamma(h_b)) < \min_{h \in [H]} \gamma(h)$.

During the decoding step of the GA-Lloyd algorithm, we may need to adjust the clusters generated in order to satisfy the minimum size constraints. If cluster $C_i$ has size smaller than $n$, then we sort the entities not in $C_i$ in the increasing order of the sum of squared regression errors, and then reassign these entities in the sorted order to cluster $C_i$ until the size of $C_i$ reaches $n$. We also skip each entity that would reduce the size of its original cluster below $n$. When there is more than one cluster with size smaller than $n$, we perform this adjustment for the smallest cluster first.

\subsection{Two-Stage Heuristic Algorithm for SKU Clustering Problem}
Due to its simplicity, two-stage heuristic algorithms are frequently employed in practice for solving the CLR problem. In the first stage, entities are partitioned according to certain approximate measures of the regression coefficients. In the second stage, regression models are built over the resultant clusters. The clustering method for the first stage is usually problem specific. 

\begin{itemize}
\item[] \textbf{Stage 1} Partition $I$ entities (SKUs) into $K$ clusters. Let $[I_k]$ be the index set of entities in cluster $k$, $k \in [K]$.
\item[] \textbf{Stage 2} For each $k \in [K]$, build a regression model using entities in $[I_k]$.
\end{itemize}

In this section, we describe our two stage heuristic algorithm for the SKU clustering problem. Recall that we are given
\begin{center}
\begin{tabular}{ll}
$\textbf{y}_i$ & weekly sales vector of SKU (entity) $i$\\
$\textbf{x}_{ij}$ & independent variable vector of SKU $i$ and dependent variable $j$
\end{tabular}
\end{center}

The first stage of our algorithm is based on hierarchical clustering and simple (one dimensional) regression. In detail, 
\begin{itemize}[noitemsep]
  \item[(1)] We carry out one regression for each SKU to get sales without promotional effects. The dependent variable for the regression is the weekly sales, and the only independent variable is the price discount. Hence, we build the following simple regression
  \begin{center}
  $\textbf{y}_i = \textbf{$\beta$}_{ij} \textbf{x}_{ij} + \textbf{r}_{ij}$,
  \end{center}
for each SKU $i$ in $[I]$ and $j = $ price discount, where $\textbf{r}_{ij}$ is the residual vector.
   Note that these regressions are one dimensional.
  \item[(2)] We construct, for each SKU, a sales vector of dimension 52 (i.e. the number of weeks in a year). The $l^{th}$ element of the vector records the mean sales without promotional effects of the $l^{th}$ week of the year, averaged over all years.  The mean sales without promotional effects are in fact $\textbf{r}_{ij}$, $j = $  price discount, in the previous step.
  \item[(3)] We calculate the correlations between any pair of sales vectors constructed in the previous step. The distance between any two SKUs is defined as one minus their corresponding correlation. That is, for any $i_1$ and $i_2$ in $[I]$, we calculate correlation $\rho_{i_1 i_2}$ between $\textbf{r}_{i_1 j}$ and $\textbf{r}_{i_2 j}$, $j = $  price discount, and define distance as $d_{i_1 i_2} = 1- \rho_{i_1 i_2}$.
  \item[(4)] We perform agglomerative hierarchical clustering over SKUs using distances generated in the previous step.  We use the maximum distance between SKUs of each cluster as the distance between any two clusters as in the complete linkage clustering \cite{Johnson:2007}.
\end{itemize}

\subsection{Sp{\"a}th Algorithm}

The algorithm in Sp{\"a}th \cite{Spath1979} is for clusterwise regression. The algorithm starts with an initial partition of the observations, then continues to move an observation to a different cluster while there is an improvement in the objective function value. The algorithm of Sp{\"a}th \cite{Spath1979} can also be used for the generalized clusterwise regression with a trivial adjustment. For generalized clusterwise regression, the only change is to move an entity (collection of observations) instead of moving an observation. Let $\displaystyle E(C_k) = \min_{\beta} \sum_{i \in C_k} \sum_{l = 1}^L \Big( y_{il} - \sum_{j = 1}^J \beta_{kj} x_{ijl} \Big)^2$. The formal procedure is in the following.
\begin{itemize}[noitemsep]
\item[] \textbf{Step 1} Start with some initial partition of entities such that $|C_k| \geq n$, $k \in [K]$, and set $i = 1$.
\item[] \textbf{Step 2} Let $i \in C_{k'}$, where $k'$ is the index of the cluster that $i$ belongs to. If $|C_{k'}| > n$ and if there exists $C_k$ with $k \neq k'$ such that $E(C_{k} \cup \{i\}) + E(C_{k'} \setminus \{i\}) < E(C_k) + E(C_{k'})$, then we pick an index $\displaystyle r = \argmin_{k \in [K], k \neq k'} E(C_{k} \cup \{i\}) + E(C_{k'} \setminus \{i\})$ and redefine $C_{k'} = C_{k'} \setminus \{i\}$ and $C_r = C_r \cup \{i\}$. In all other cases, set $i := i+1$ and return to Step 2.
\item[] \textbf{Step 3} Repeat Step 2 until there is no improvement in Step 2 for $I$ consecutive times.
\end{itemize}
In Step 1, each entity is assigned to a cluster based on a uniformly distributed random number. In Step 2, we move entity $i$ to the cluster that reduces the objective function the most. Step 2 is repeated until no entry can be moved with a reduction of the objective function value. In the rest of the paper, we denote this algorithm as Sp{\"a}th.

\section{Numerical Experiments}\label{Sec:Experiments}
In this section, we examine the performance of CG (Algorithm \ref{alg:exactSCG}), CG Heuristic (Algorithm \ref{alg:CGHeur}), GA-Lloyd (Algorithm \ref{alg:GALloyd}), Sp{\"a}th and the two-stage algorithms on the SKU clustering problem according to its seasonal effects. The regression model for this problem has the following form:
\begin{equation*}
f_0(\text{weekly sales}) = f_1(\text{promotional predictors}) + f_2(\text{seasonal predictors}),
\end{equation*}
where the seasonal effects are modeled by 52 dummy variables, one for each week. There are three types of data sets we used in the experiment.
\begin{enumerate}
\item Real-world data: We use real-world data from a large retail chain. We omit the exact form of the regression model due to confidentiality. It includes more than two years of aggregated sales and promotional data of the entire chain. The products within this chain are grouped into subcategories for purchasing purposes. However, it is assumed that the products within the same subcategory have different seasonal patterns with regard to promotions. We tested our algorithms on two representative subcategories from the data, a smaller subcategory ``Cream'' and one of the largest subcategories ``Medicines.'' Both subcategories have more than one seasonal pattern. Each SKU in these subcategories has at least 52 weeks of data and at most 129 weeks of data from year 2006 to 2008. 
\item Synthetic data type 1: We generate random instances that have similar patterns as the real-world data. The promotional predictor includes percentage of discount and the seasonal predictor captures week index. Therefore, the data set includes weekly sales, percentage of discount, and the week index. Each entity has a year of records (52 weeks). For each $I \in \{15,20,25,50,100,150,200\}$, we generate 10 random instances, which results into 7 different size data sets of total 70 instances. The detailed instance generation procedure is available in Appendix \ref{appendix_instance_generation_1} and the data set is available at a web site \footnote[1]{\url{dynresmanagement.com/public_data/sku_clustering_random_data.zip}}. 
\item Synthetic data type 2: We generate random instances based on the procedure for synthetic data type 1. The difference is that we are given the target number of cluster and entities are assigned to a cluster. For this reason, we have a target solution for each instance that is expected to be good. The target solution can be used to evaluate heuristic algorithms. However, in our experiment, we observe that most of the proposed algorithms give better solutions than the target solution in terms of the objective function value. Hence, in this paper, we did not use the target solution for the evaluation of the algorithms. The structure and size of the data are the same as in the synthetic data type 1. The generation procedure is available in Appendix \ref{appendix_instance_generation_2} and the data set is available at a web site$^1$.
\end{enumerate}

Note that the real-world data and the implementation of the two stage algorithm had to be destroyed before the publication. For this reason, the implementations used for real-world and synthetic data are slightly different. For the same reason, the computational environments used and performance measures are different for real-world and synthetic data. In Table \ref{tab:data_and_algo}, we summarize which algorithms are used for each data set.

\begin{table}[h]
\centering
\small
\begin{tabular}{|c|c|c|c|c|c|}
\hline
 & CG & CGH & GA-Lloyd & Two Stage & Sp{\"a}th \\ \hline
Real-World & O & O& O& O & \\
Synthetic & O& O& O& & O \\
\hline
\end{tabular}
\caption{Algorithms used for the data sets} \label{tab:data_and_algo}
\end{table}

The following computational environments were used for real-world and synthetic data sets.
\begin{enumerate}
\item Real-world data: All the algorithms except the two-stage algorithm were implemented in Java 1.6 with CPLEX 11 as the mathematical programming solver. The ``lm'' function in R 2.8 is used to perform regressions for the GA-Lloyd algorithm. The two-stage algorithm is implemented in R 2.8, and the ``hclust'' function for hierarchical clustering is employed to perform clustering. All numerical experiments were performed on a 64-bit server with a multi-core Intel Xeon 2 GHz CPU and 10 GB of RAM.
\item Synthetic data types 1 and 2: All the algorithms were implemented in Java 1.7 with CPLEX 12.5 and R 2.8. The experiments were performed on a 64-bit server with a multi-core Intel Xeon 2.8 GHz CPU and 32 GB of RAM.
\end{enumerate}

%Implementation details
%??? MODIFIED BY CONRADO
%BEFORE
%The convergence of column generation is slow due to degeneracy. We apply the scheme discussed in \citet{duMerle1999} to stabilize the column generation algorithm and speed up convergence.
Note also that, for all experiments in Section 4, specific values for the following parameters of the GA-Lloyd algorithm were selected based on a sensitivity analysis: $H=10$ (population size) and $p = 0.01$ (permutation probability). Furthermore, as a termination criterion for the algorithm we specify that the number of consecutive iterations with no improvement has to reach 50. In the experiments, we use a very large big M relative to the residuals to ensure optimality. On average, $M$ is 3000 times larger than the average of absolute residuals and $M$ is 30 times the average of the response variable.

\subsection{Real-World Data}

\subsubsection{Comparison of the Heuristic Algorithms}
\label{subsubsection_compare_heur_real}

Based on preliminary computational studies of the column generation algorithm and the results shown in Section~\ref{subsub:TimeStudyColumnGeneration}, we observed that the algorithm does not scale well when applied to large size instances. Hence, the heuristic algorithms are employed to cluster SKUs for the ``Cream'' and ``Medicine'' subcategories. 

%Modified by Conrado
In this section we compare the performance of the GA-Lloyd, CG Heuristic, and two-stage algorithms with $K \in \{2,3,4,5\}$ and $n=3$ in terms of solution time and quality. As the initial important remark, we observe that the GA-Lloyd algorithm performs the best while providing a good balance between time and quality. In the experiments, an instance denoted by ``$I\_K$'' means that we divided $I$ SKUs into $K$ clusters. For example, an instance denoted by ``66\_2'' implies that we divided 66 SKUs into two clusters. The stopping criteria for the CG Heuristic algorithm are: (a) an optimal solution is found by the column generation with $R$ groups of entities, and (b) a time limit of 10 hours is reached after the last pricing problem. For subcategory ``Cream,'' the CG Heuristic algorithm applied stopping criterion (a), whereas for the subcategory ``Medicine,'' it stopped due to (b). 

In this section, for any two algorithms $algo1$ and $algo2$, we use relative improvement (RI\%) by $algo2$ from $algo1$, which is defined by
\begin{center}
RI($algo1,algo2$) = $\frac{\hbox{SSE(\textit{algo1})} - \hbox{SSE(\textit{algo2})}}{\hbox{SSE(\textit{algo2})}},$
\end{center}
where the value of SSE(algorithm) is the sum of the regression errors across all clusters. Note that RI($algo1,algo2$) $>$ 0 means that $algo2$ generates a better solution and RI($algo1,algo2$) $<$ 0 implies superiority of $algo1$.

\begin{figure}[ht]
\center
  \includegraphics[scale=0.7]{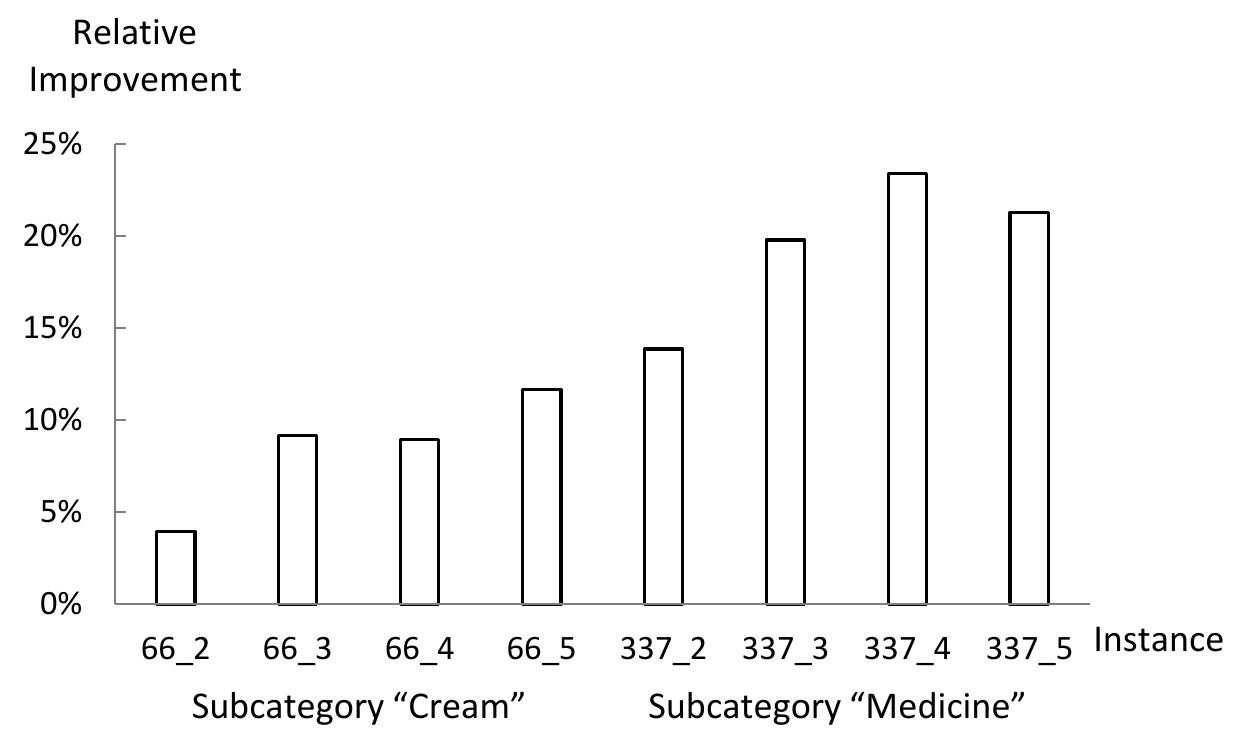}\\
  \caption{Relative Improvement of GA-Lloyd over Two-Stage for Real-World Data}
  \label{Fig:GA-LloydVSTwoStage}
\end{figure}

%We compare the GA-Lloyd algorithm with the two-stage algorithm. We divide the subcategories into two to five clusters with a minimum size of three. 
%\textbf{Confusing!} 
First, we compare the GA-Lloyd algorithm with the two-stage algorithm. Figure \ref{Fig:GA-LloydVSTwoStage} shows RI(Two Stage,GA-Lloyd), which is the relative improvement of the GA-Lloyd algorithm over the two-stage algorithm. We observe that a substantial improvement is achieved by the GA-Lloyd algorithm over the two-stage heuristic. 

In terms of the running time, we observe that the two-stage algorithm outperforms the GA-Lloyd algorithm while converging within one and five minutes, respectively, for the ``Cream'' and ``Medicine" subcategories, whereas the GA-Lloyd algorithm took roughly 20 minutes and one hour for the corresponding subcategories. In practice, however, run times of one-hour are acceptable for the SKU clustering problem according to our retail partner.

\begin{figure}[ht]
\center
  \includegraphics[scale=0.7]{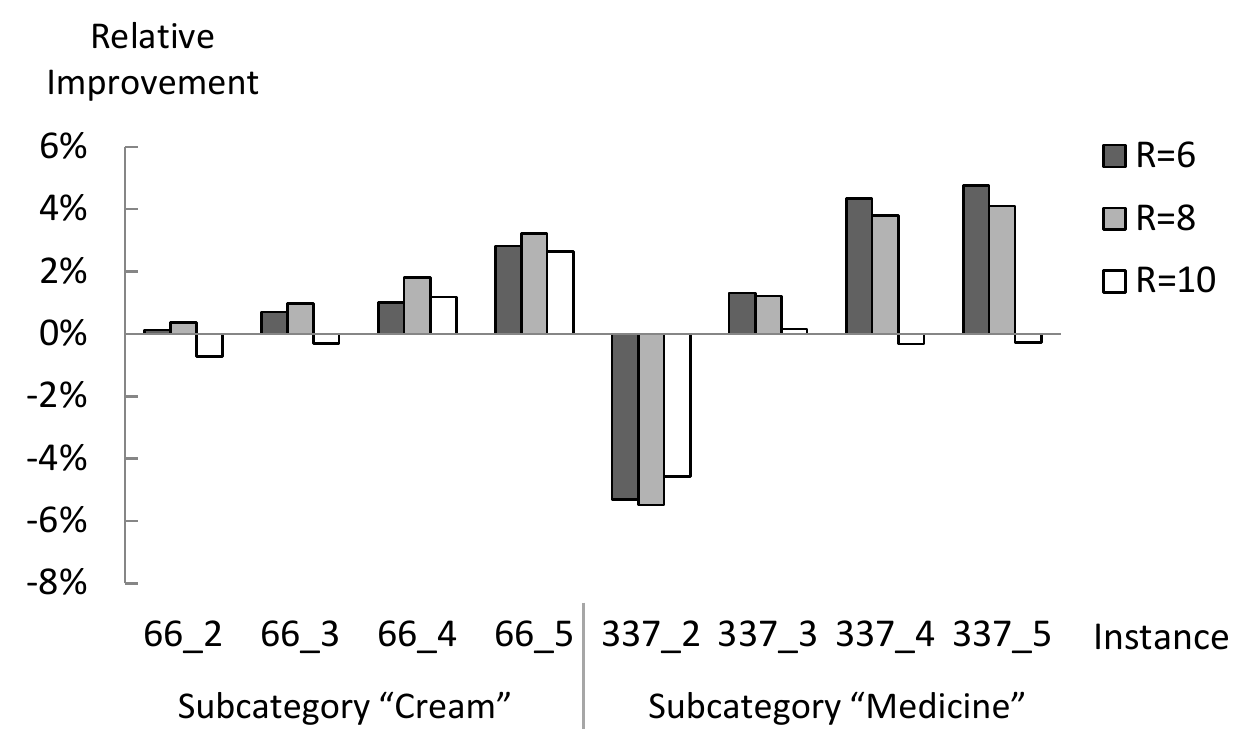}\\
  \caption{Relative Improvement of GA-Lloyd over CG Heuristic for Real-World Data}
  \label{Fig:CGHeuristicVsGA}
\end{figure}

\begin{figure}[ht]
     \begin{center}
        \subfigure[Subcategory ``Cream'']{
            \label{Fig:Time686}
            \includegraphics[scale=0.27]{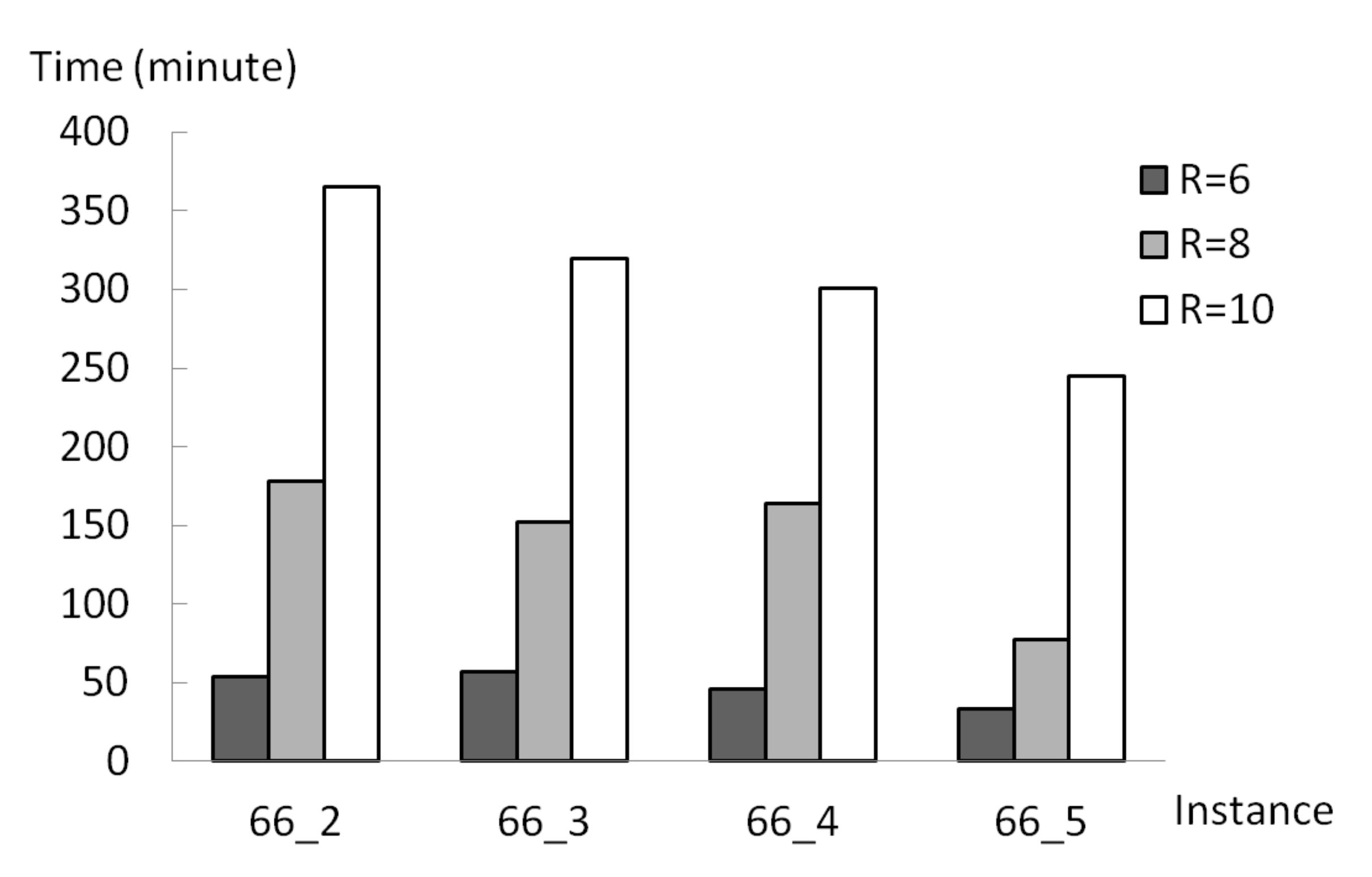}
        }
        \subfigure[Subcategory ``Medicine'']{%
           \label{Fig:Time344}
           \includegraphics[scale=0.27]{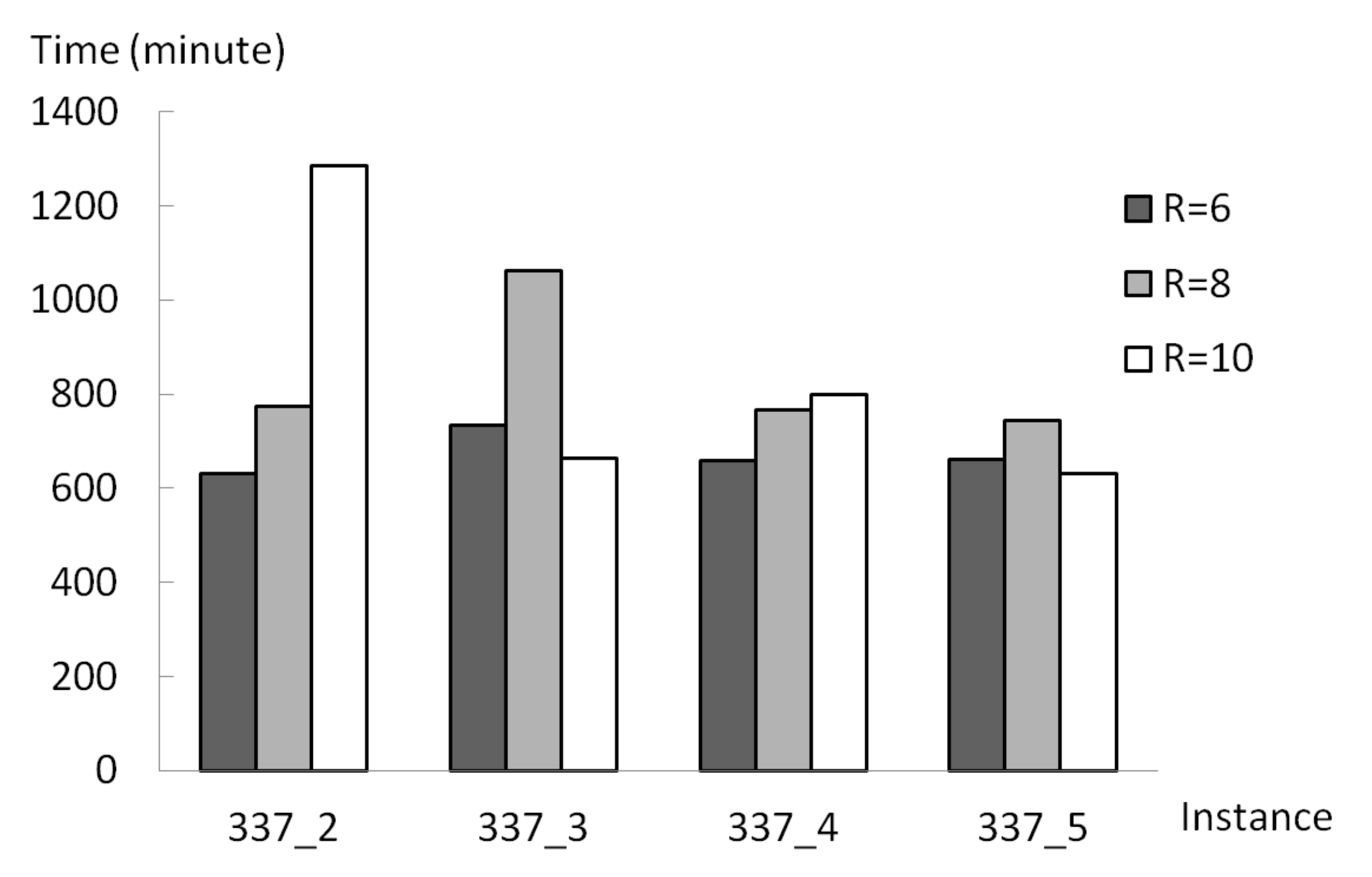}
        }
    \end{center}
    \caption{CG Heuristic Running Time for Real-World Data}
   \label{Fig:CGHeuristicRunningTime}
\end{figure}

%Modified by Conrado 
Figure \ref{Fig:CGHeuristicVsGA} shows RI(CG Heuristic,GA-Lloyd) with $R = \{6,8,10\}$ for the CG Heuristic algorithm. We observe that the resulting $\mbox{CostDiff}$ values between the GA-Lloyd and CG Heuristic algorithms are not significant, all within 6\%. The CG Heuristic algorithm with 
%$R$ equal to 10 
$R = 10$ 
generates better solutions than the GA-Lloyd algorithm for five out of eight instances. Among the five instances, in four of them the improvement is barely noticeable (``337\_2'' is the only case with more pronounced improvement). However, its running time, four to six hours for subcategory ``Cream'' and 10 hours for subcategory ``Medicine'', is much longer than that of the GA-Lloyd algorithm, as illustrated in Figure \ref{Fig:CGHeuristicRunningTime}. In summary, it is recommended to select largest $R$ within affordable time limit for solving the problem.

%Modified by Conrado
From Figure \ref{Fig:CGHeuristicVsGA}, we also observe that the GA-Lloyd algorithm performs better than the CG Heuristic algorithm with 
% $R$ equal to 6 and 8 
 $R=\{6, 8\}$ 
 for all but one instance. In addition, the GA-Lloyd algorithm also outperforms the CG Heuristic algorithm in terms of computational times for all these instances, 
as shown in Figure \ref{Fig:CGHeuristicRunningTime}. When comparing the solutions of the CG Heuristic algorithm with $R = 6$ and 8  
%$R$ equal to 6 and 8 
for subcategory ``Cream'', Figure \ref{Fig:CGHeuristicVsGA} shows that a higher value of R does not necessarily imply a better solution. 

\subsubsection{Seasonal Patterns Identified by GA-Lloyd}
\label{subsec_compare_seasonality}

Based on the superiority of the GA-Lloyd algorithm over its counterparts shown in the previous section, in this section we present numerical results of the seasonal patterns identified by this algorithm when applied to the largest subcategory, ``Medicine". 

\begin{figure}[ht]
    \begin{center}
        \includegraphics[width = 0.9\linewidth]{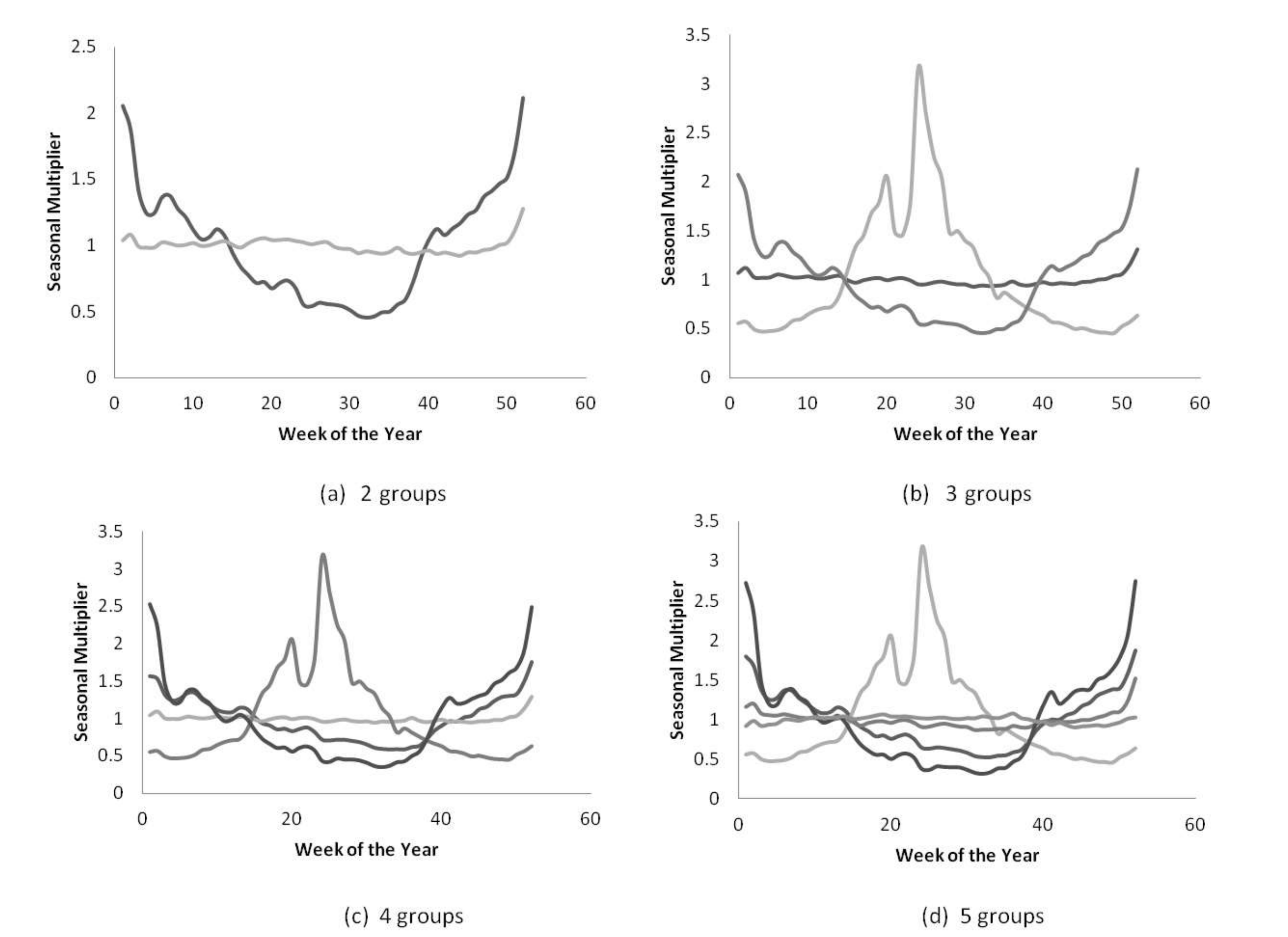}
    \end{center}
    \caption{Subcategory ``Medicine'' Seasonal Multipliers for Real-World Data}\label{Fig:SeasonalMedicine}
\end{figure}

Figure \ref{Fig:SeasonalMedicine} shows the seasonal multipliers obtained from the GA-Lloyd algorithm for each cluster when dividing the subcategory ``Medicine'' into 2, 3, 4 and 5 clusters. From the figure, we observe distinct seasonal patterns: (1) U-shaped curve, (2) inverted V-shaped, and (3) flat. Observe that all of the three seasonal patterns have been found when dividing SKUs into three clusters. This observation indicates that it is not necessary to further divide them into four or five clusters since some seasonal patterns look similar. The three clusters of SKUs represent medicines that intuitively have such different seasonal patterns: one corresponding to medicines (such as for cold and flu) that sell more in the winter, one corresponding to medicines (such as for bug repellents and sunburns) with uplift in the summer, and one corresponding to medicines (such as for diarrhea and constipation) with stable sales year around.

\subsubsection{Optimality Gap of GA-Lloyd and CG Heuristic}\label{subsubsection:OptimalityGap}

In this section, we benchmark the performance of the GA-Lloyd and CG Heuristic algorithms by comparing against the CG algorithm, which is an optimal algorithm. In order to measure the performance, we calculate 
\begin{equation}
\mbox{OptGap} = \left[\frac{\hbox{SSE(\textit{algo})} - \hbox{SSE(Column Generation)}}{\hbox{SSE(Column Generation)}}\right], \label{eqn_optgap}
\end{equation}
where \textit{algo} $\in \{$ CG Heuristic, GA-Lloyd $\}$. For the real-world data, note that since the column generation algorithm did not cluster SKUs for the ``Cream'' and ``Medicine'' subcategories within a reasonable computational time due to their large sizes, we construct smaller instances with 15 and 20 SKUs that are randomly chosen from the large subcategory ``Medicines.'' For these instances, we test the algorithms for parameters $K \in \{2,3,4\}$ and $n = 2$. We also studied the instance that divides the subcategory ``Cream'' with 66 SKUs into two clusters with minimum cluster size $n = 3$, which we refer to as $66_2$ instance in this section. The column generation exact algorithm were executed with 10 hours of time limit for the $66_2$ instance.

\begin{figure}[ht]
     \begin{center}
        \subfigure[GA-Lloyd Algorithm]{
            \includegraphics[scale=0.6]{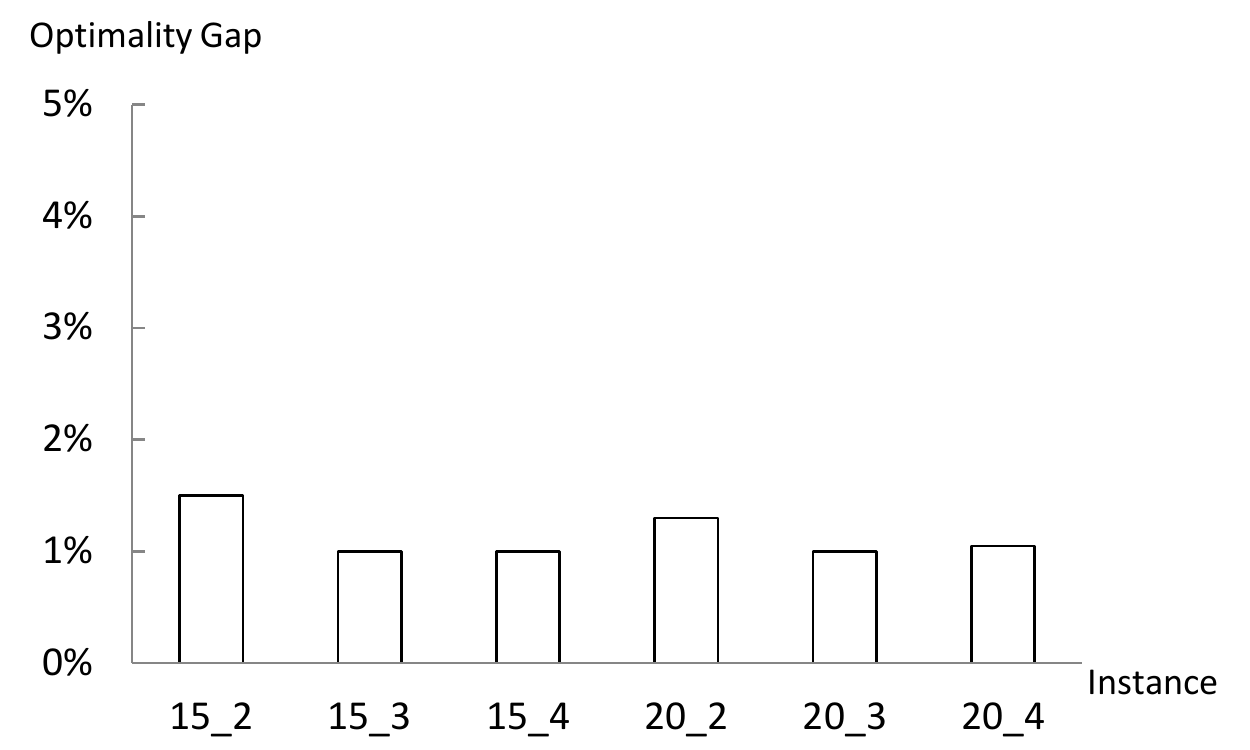} \label{Fig:GAOptimalityGap}
        }
        \subfigure[CG Heuristic Algorithm]{%
           \includegraphics[scale=0.6]{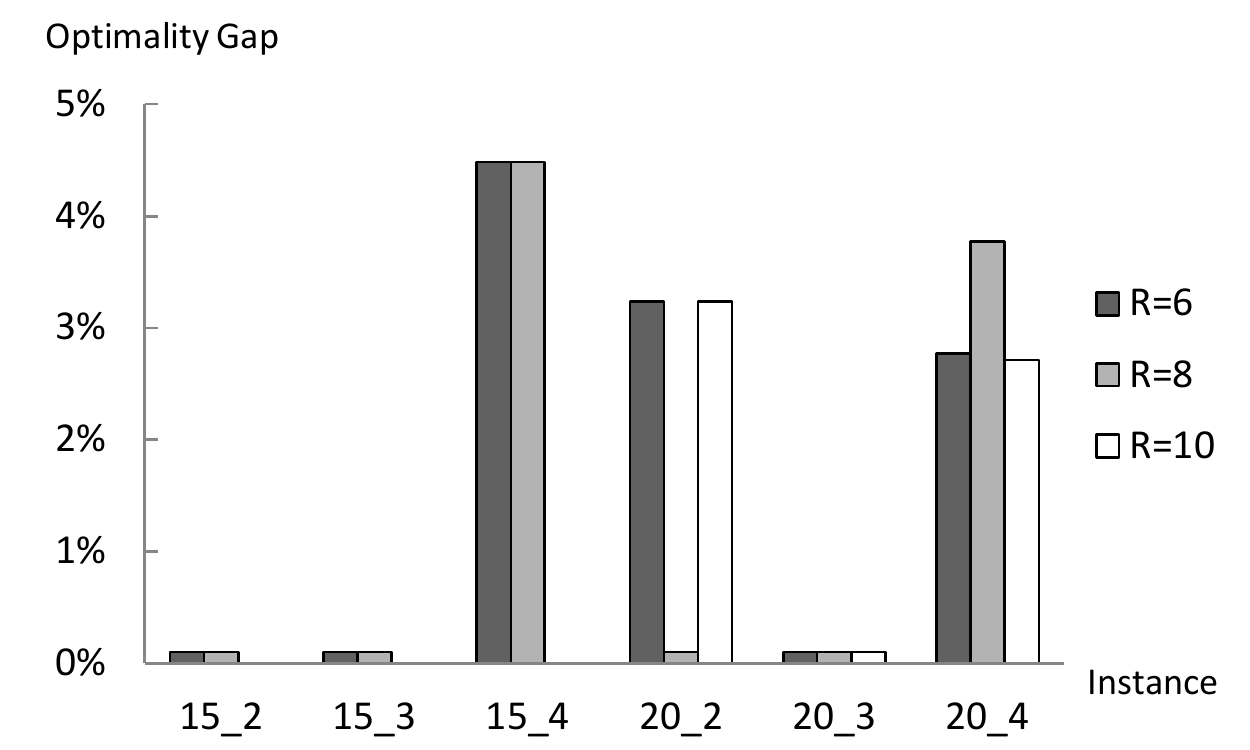} \label{Fig:CGHeuristicOptimalityGap}
        }
    \end{center}
    \caption{Optimality Gap of GA-Lloyd and CG Heuristic for Real-World Data}
\end{figure}

Figure \ref{Fig:GAOptimalityGap} shows the optimality gap values of the GA-Lloyd algorithm. We observe that the GA-Lloyd algorithm achieves close to optimal solutions, with optimality gaps less than 2\%. In addition, the GA-Lloyd algorithm finishes within five minutes for these smaller instances. 

For the $66_2$ instance, the solution obtained from the column generation exact algorithm after 10 hours of running time is only 1.47\% better than the GA-Lloyd solution, which is obtained in less than 20 minutes. This case is not shown in the figure.

Figure \ref{Fig:CGHeuristicOptimalityGap} shows the optimality gap values of the CG Heuristic algorithm for different values of $R$. We observe that the CG Heuristic algorithm also finds close to optimal solutions, with optimality gaps less than 5\%. In addition, the CG Heuristic algorithm finishes within five minutes for the instances of size 15, and within 20 minutes for the instances of size 20. By comparing the solutions for the instance ``20\_2'' for $R$ equal 8 and 10, we again find that a larger $R$ does not necessary generate a better solution. 

For the $66_2$ instance, the solution obtained from column generation exact algorithm after 10 hours of running time is only 0.13\%, 1.86\%, 0.75\% better than the CG Heuristic solution with $R$ equal to 6, 8, and 10, respectively. The corresponding running times of the CG Heuristic algorithm are 54, 178, and 365 minutes, respectively. The performance for the 66 SKUs is not depicted graphically.

%Modified by Conrado
The average gap in Figure \ref{Fig:GAOptimalityGap} is 1.14\%, while in Figure \ref{Fig:CGHeuristicOptimalityGap} for $R = 10$ it is 1.01\%. This indicates that for smaller instances the CG Heuristic algorithm outperforms the GA-Lloyd algorithm if the objective value is the only performance indicator and the computational time is limited to 10 hours. On the contrary, Figure \ref{Fig:CGHeuristicVsGA} indicates clearly that 
%for larger instances GA-Lloyd is the winner.
the GA-Lloyd algorithm suits better for larger instances. This implies that whenever a strict run time limit is imposed, the GA-Lloyd algorithm is very likely to outperform its counterparts for most of the instances.

\subsubsection{Time Study of the Column Generation Algorithm}\label{subsub:TimeStudyColumnGeneration}

\begin{figure}[ht]
     \begin{center}
        \subfigure[Sum of Squared Errors]{
            \label{Fig:SSE}
            \includegraphics[scale=0.6]{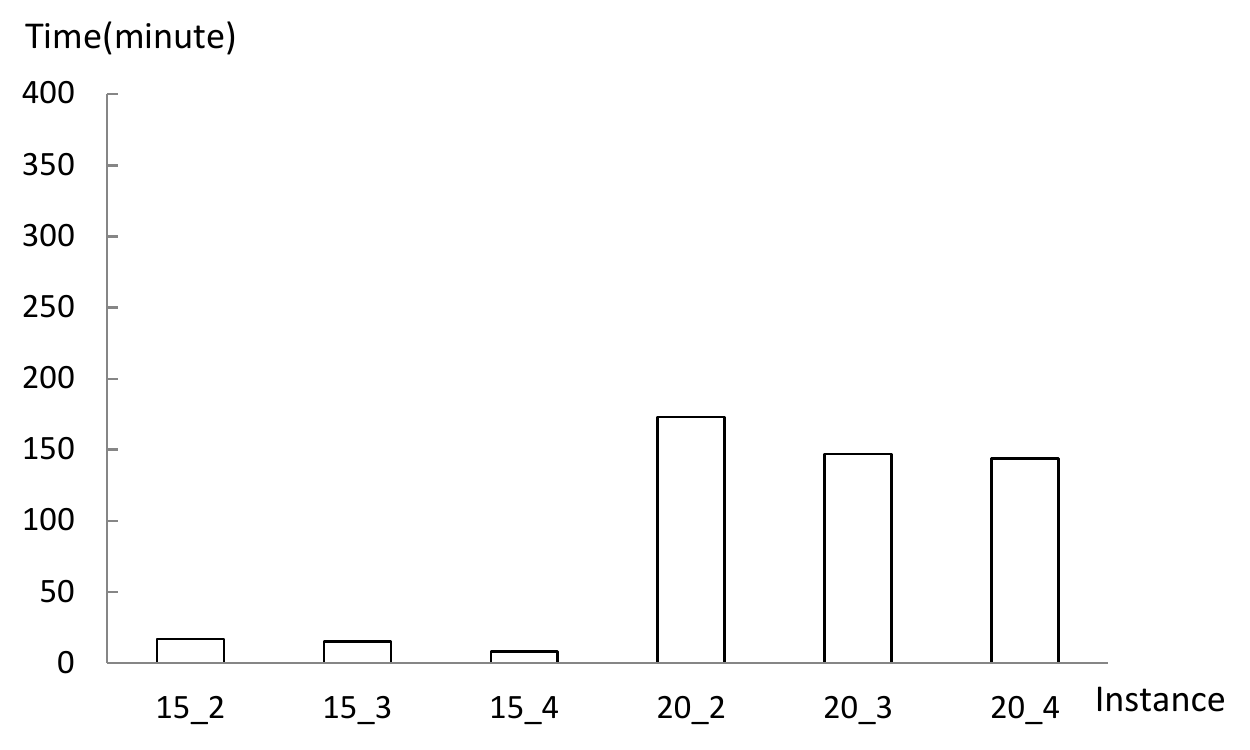}
        }
        \subfigure[Sum of Absolute Errors]{%
           \label{Fig:SAE}
           \includegraphics[scale=0.6]{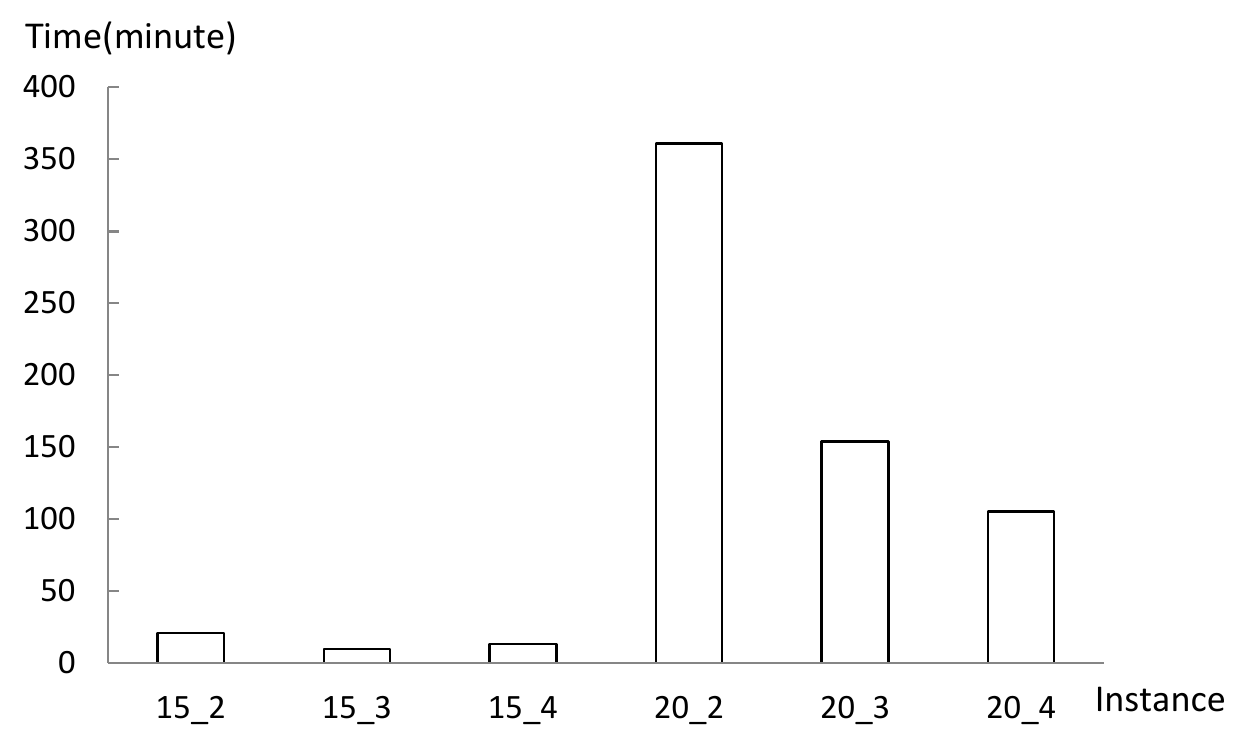}
        }
    \end{center}
    \caption{Column Generation Running Time for Real-World Data}
   \label{Fig:CGRunningTime}
\end{figure}

The performance of the Column Generation algorithm (Algorithm \ref{alg:exactSCG}) is assessed on a set of computational experiments conducted on instances with 15, 20 and 66 SKUs. These instances are chosen in a similar way as in the previous section. Figure \ref{Fig:SSE} presents the running time of Algorithm \ref{alg:exactSCG} for specific numbers of SKUs and clusters to divide in. 
%We observe that the solution time increases quickly as the number of SKUs to cluster increases. For example, w
We can observe that it takes roughly 3 hours to divide 20 SKUs into two clusters.

For the instance with 66 SKUs and two clusters, we are unable to get an optimal solution after 10 hours. When comparing to the lower bound obtained by solving the linear relaxation of the mixed integer formulation \eqref{Eq:MixedIntegerQuadratic}--\eqref{constraint:quadratic:minimumSize}, the solution gotten from column generation after 10 hours of running time is 38.23\% larger than the lower bound. However, we suspect this solution is very close to the optimal one because we observe that the minimum reduced cost of the master problem is close to zero.

We also study another version of the generalized CLR problem with the sum of absolute errors as the objective to examine whether the difficulty in solving the pricing problem is due to the nonlinearity of the objective function in the pricing problem \eqref{Eq:Pricing}-\eqref{constraint:pricing:MinSize}. More specifically, we change the objective function in the pricing problem to $\sum_{i = 1}^I\sum_{l = 1}^{L}t_{il}-\sum_{i = 1}^I{\pi_iz_{i}}$. Figure \ref{Fig:SAE} presents the running time when the objective function for the CLR problem is the sum of absolute errors. From this figure, we observe that the running time also increases dramatically as the number of SKUs increases. It also takes hours to solve the instances with 20 SKUs. Therefore, we believe the nonlinear objective of the pricing problem is not the complicating factor that greatly drives up the computation time. These running times are higher due to a larger number of iterations resulting from degeneracy of LP solutions (the per-iteration time is lower). 

\subsection{Synthetic Data}

\subsubsection{Optimality Gap of the Heuristic Algorithms}

\begin{figure}[ht]
     \begin{center}
             \subfigure[Optimality Gap]{%
        \label{fg_exp_type1_small_gap}
           \includegraphics[scale=0.6]{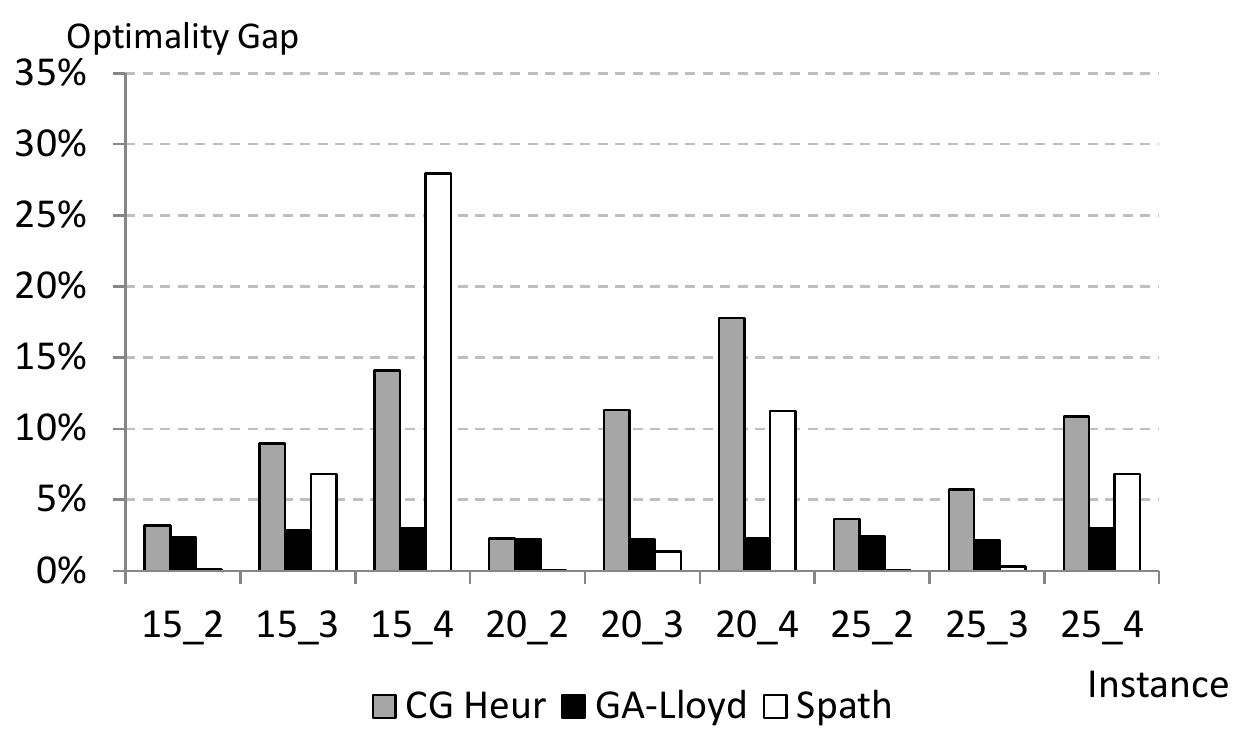}
        }
        \subfigure[Running Time]{
        \label{fg_exp_type1_small_time}
            \includegraphics[scale=0.6]{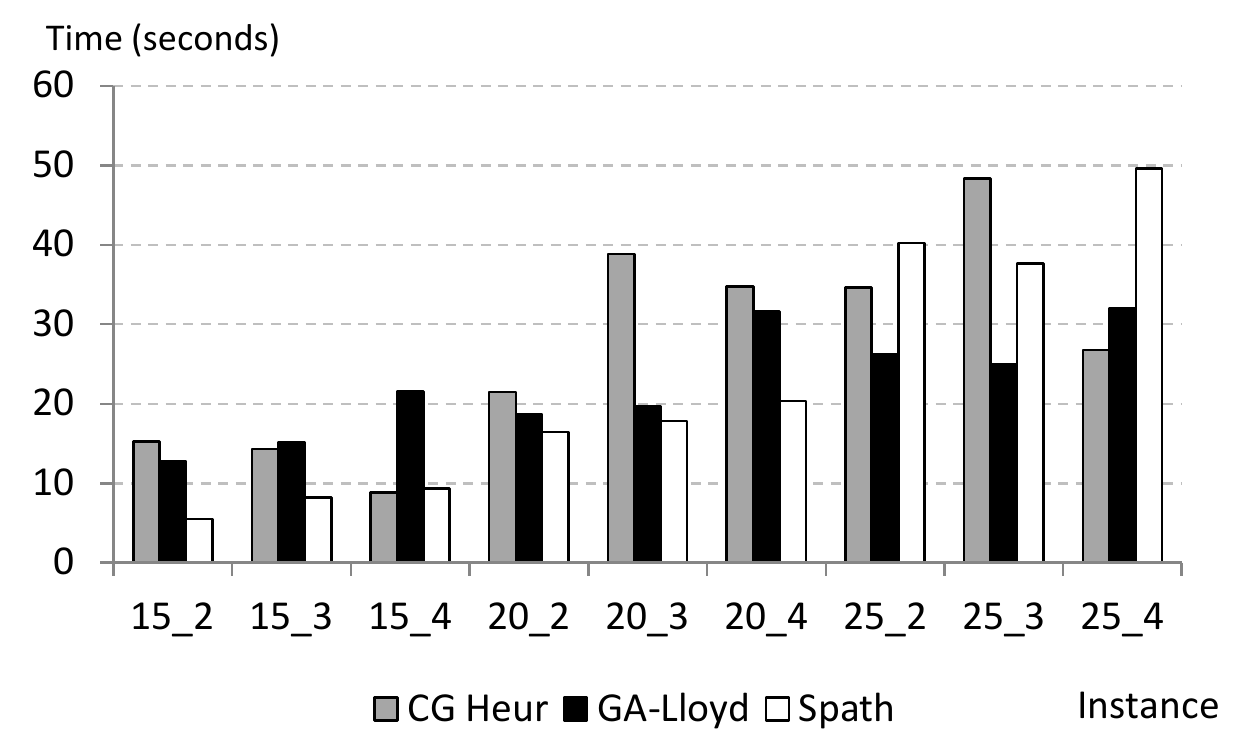}
        }
    \end{center}
    \caption{Comparison of the Heuristic Algorithms for Synthetic Data Type 1 } \label{fg_exp_type1_small}
\end{figure}

\begin{figure}[ht]
     \begin{center}
        \subfigure[Optimality Gap]{%
        \label{fg_exp_type2_small_gap}
           \includegraphics[scale=0.6]{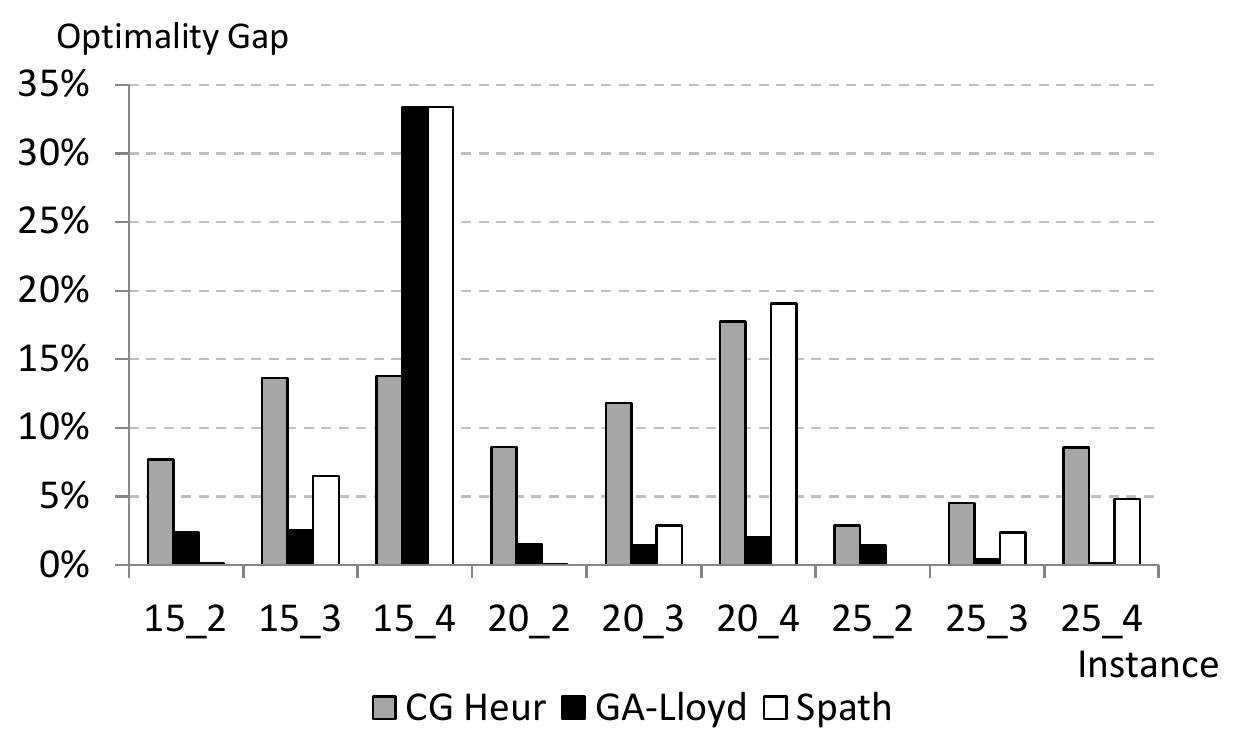}
        }
        \subfigure[Running Time]{
        \label{fg_exp_type2_small_time}
            \includegraphics[scale=0.6]{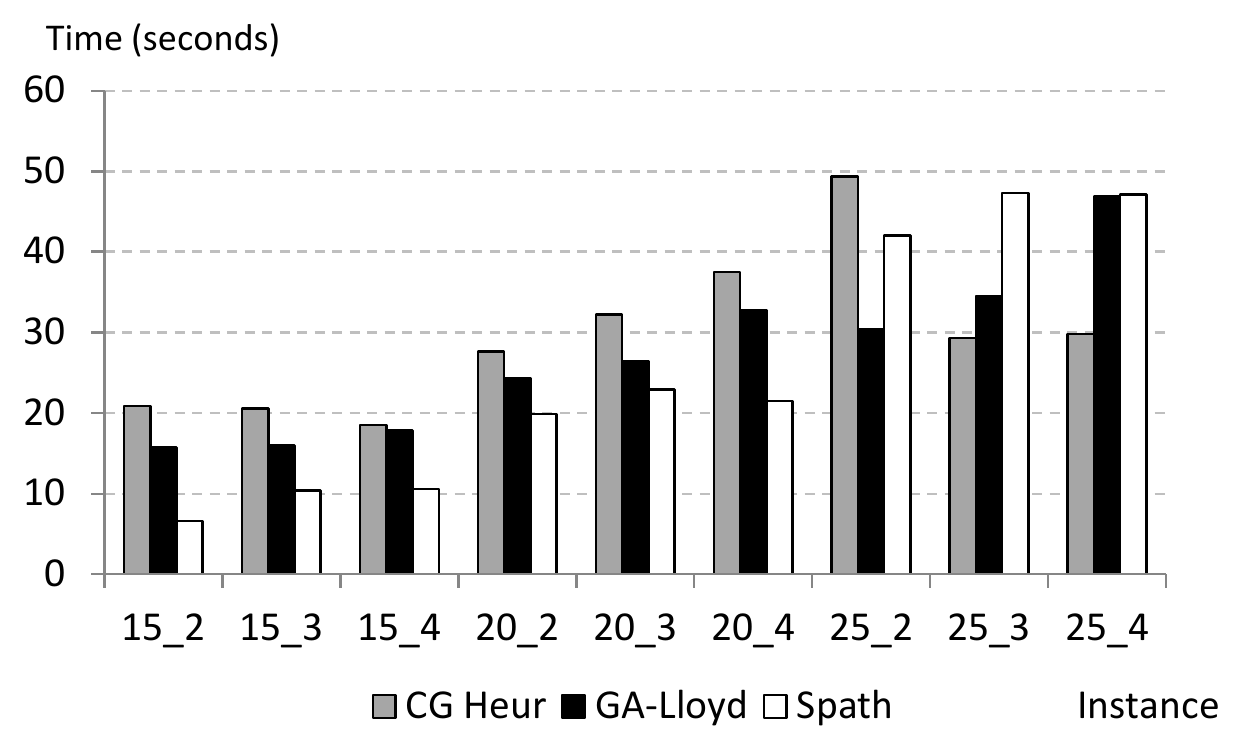}
        }
    \end{center}
    \caption{Comparison of the Heuristic Algorithms for Synthetic Data Type 2 } \label{fg_exp_type2_small}
\end{figure}

In this section, we test synthetic instances (types 1 and 2) with $I \in \{15,20,25\}$ for parameters $K \in \{2,3,4\}$ and $n = 2$. These instances are (except for type 2 with $I=25$) of the size that can be optimally solved within reasonable amount of time. We test the performance of the CG Heuristic, GA-Lloyd, and Sp{\"a}th algorithms by comparing against the solution of CG for the small synthetic instances with $I \in \{15,20,25\}$ and $K \in \{2,3,4\}$, where the optimality gap defined in \eqref{eqn_optgap} is used. We only execute the CG Heuristic algorithm with $R = 8$. For type 2, we did not run CG algorithm for instances with $I=25$ due to excessive computation time. Instead, for instances with $I=25$, we consider the best objective function value among all algorithms as optimal for calculating the gap. We present the result in Figures \ref{fg_exp_type1_small} and \ref{fg_exp_type2_small} for types 1 and 2 data, respectively. In Figures \ref{fg_exp_type1_small_time} and \ref{fg_exp_type2_small_time}, we observe that the running times of the algorithms are of the same magnitude. We also observe that the running time of the CG Heuristic algorithm tends to decrease in $K$ and the running time of the GA-Lloyd algorithm tends to increase in $K$. The optimality gap in Figure \ref{fg_exp_type1_small_gap} shows that Sp{\"a}th performs best for $K = 2$ but the optimality gap drastically increases in increasing $K$. The GA-Lloyd algorithm gives less than 5\% optimality gap for all data sizes. Figure \ref{fg_exp_type2_small_gap} also shows the same pattern except for one data set with $I = 15$ and $K = 4$. Hence, we conclude that the performance of GA-Lloyd is stable and good. The CG algorithm does not perform best for these small instances.

\subsubsection{Comparison of the Heuristic Algorithms}

\begin{figure}[ht]
     \begin{center}
        \subfigure[Gap from the best objective]{%
        \label{fg_exp_type1_large_gap}
           \includegraphics[scale=0.6]{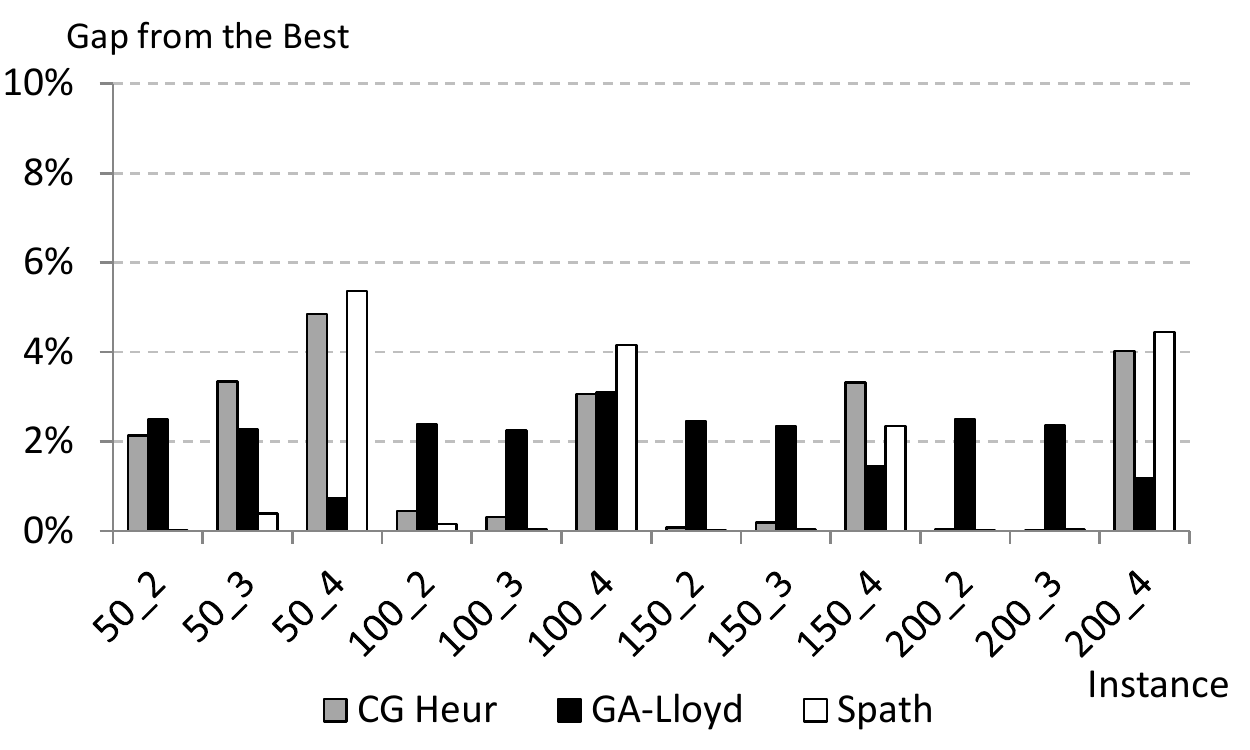}
        }
        \subfigure[Running Time]{
        \label{fg_exp_type1_large_time}
            \includegraphics[scale=0.6]{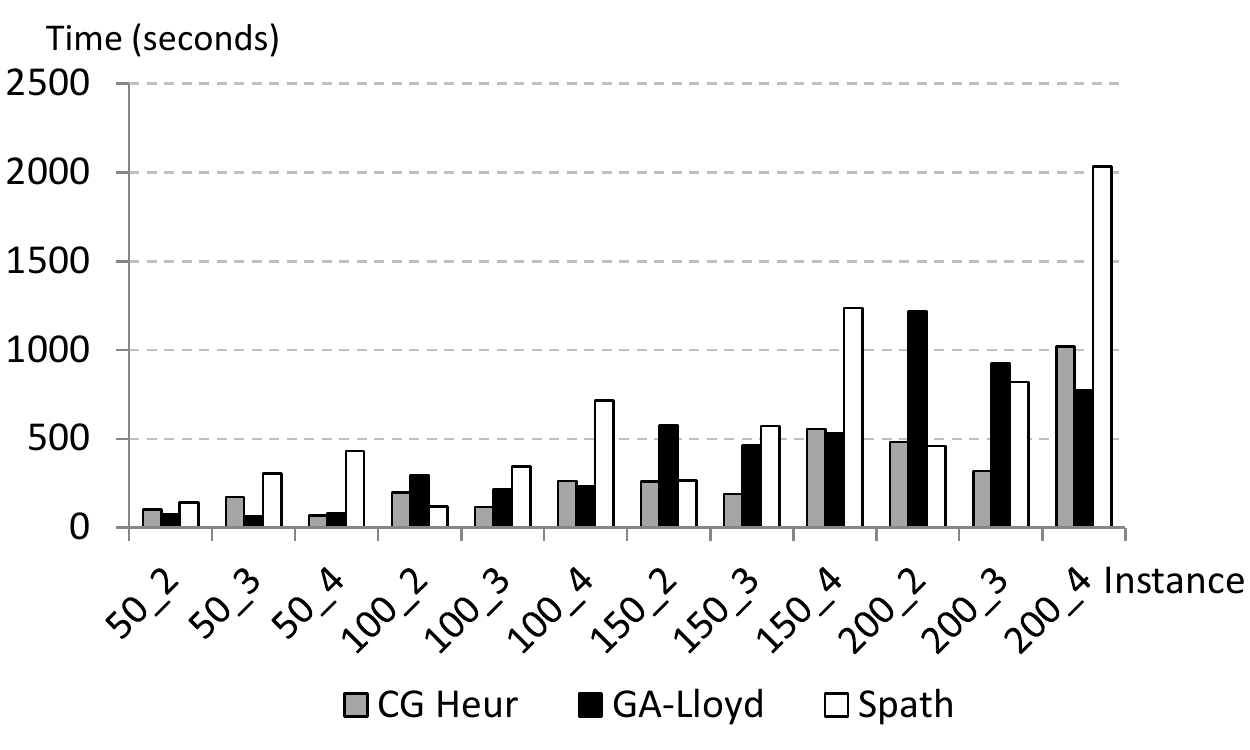}
        }
    \end{center}
    \caption{Comparison of the Heuristic Algorithm for Synthetic Data Type 1 } \label{fg_exp_type1_large}
\end{figure}

\begin{figure}[ht]
     \begin{center}
        \subfigure[Gap from the best objective]{%
        \label{fg_exp_type2_large_gap}
           \includegraphics[scale=0.6]{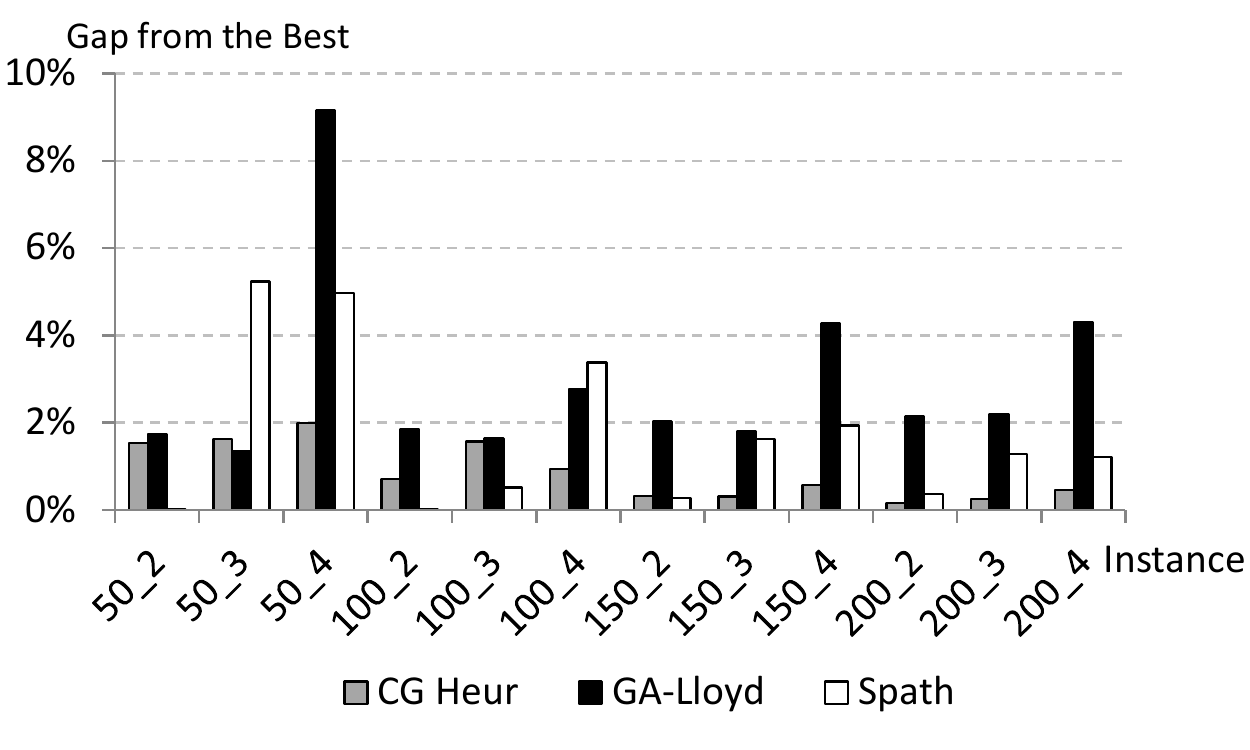}
        }
        \subfigure[Running Time]{
        \label{fg_exp_type2_large_time}
            \includegraphics[scale=0.6]{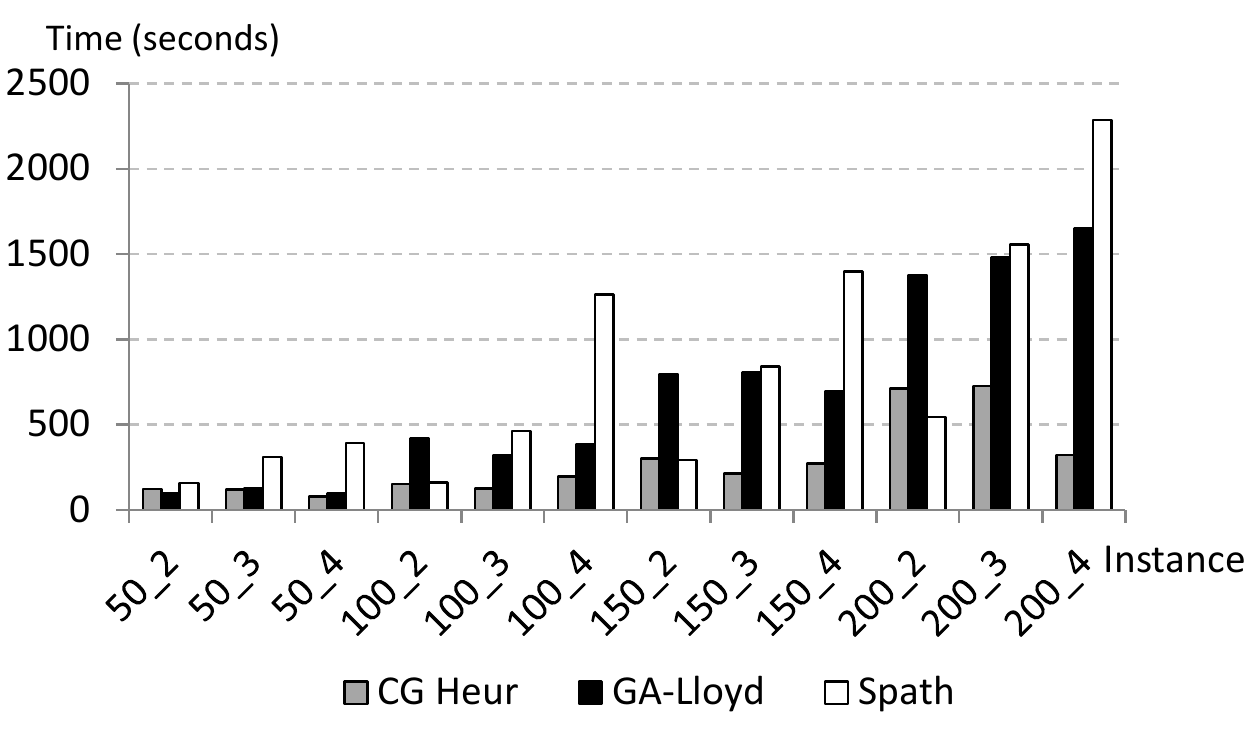}
        }
    \end{center}
    \caption{Comparison of the Heuristic Algorithm for Synthetic Data Type 2 } \label{fg_exp_type2_large}
\end{figure}

In this section, we compare the performance of the CG Heuristic, GA-Lloyd, and Sp{\"a}th algorithms with $K \in \{2,3,4\}$ and $n = 3$ for the synthetic instances with $I \in \{50,100,150,200\}$. Due to its excessive computational time, we did not run the CG algorithm. Instead, we use 
\begin{equation}
\mbox{Gap} = \frac{\mbox{SSE(\textit{algo})} - \mbox{min} \big\{ \mbox{SSE(CG Heuristic)}, \mbox{SSE(GA-Lloyd)}, \mbox{SSE(Sp{\"a}th)} \big\}}{\mbox{min} \big\{ \mbox{SSE(CG Heuristic)}, \mbox{SSE(GA-Lloyd)}, \mbox{SSE(Sp{\"a}th)}\big\}}. \label{gap_formula}
\end{equation}
We execute the CG Heuristic algorithm with $R = 8$. The result is presented in Figures \ref{fg_exp_type1_large} and \ref{fg_exp_type2_large} for types 1 and 2 data, respectively. From Figures \ref{fg_exp_type1_large_gap} and \ref{fg_exp_type2_large_gap}, we observe that Sp{\"a}th generally performs the best for small $K$. However, the gap drastically increases as $K$ increases, and Sp{\"a}th is recommended to be used for small $K$. The performance of the CG Heuristic is competitive for type 1 data and is the best among all for type 2 data. The performance of the GA-Lloyd is not competitive for both types of data sets. The computation times in Figures \ref{fg_exp_type1_large_time} and \ref{fg_exp_type2_large_time} show that the scalability of the algorithms is in the order of CG Heuristic, GA-Lloyd, and Sp{\"a}th. The running time of Sp{\"a}th drastically increases in problem size.

\subsubsection{Comparison of the Heuristic Algorithms Over Time}
In this section, we compare the performance of the CG Heuristic, GA-Lloyd, and Sp{\"a}th algorithms over time with $K \in \{2,3,4\}$ and $n = 3$ for the type 1 synthetic instances with $I \in \{100,150,200\}$. We present how the best solution of each algorithm is updated over time, while the result in Section 4.2.2 is based on the solutions at the termination. 

\begin{figure}[ht]
\center
  \includegraphics[scale=0.3]{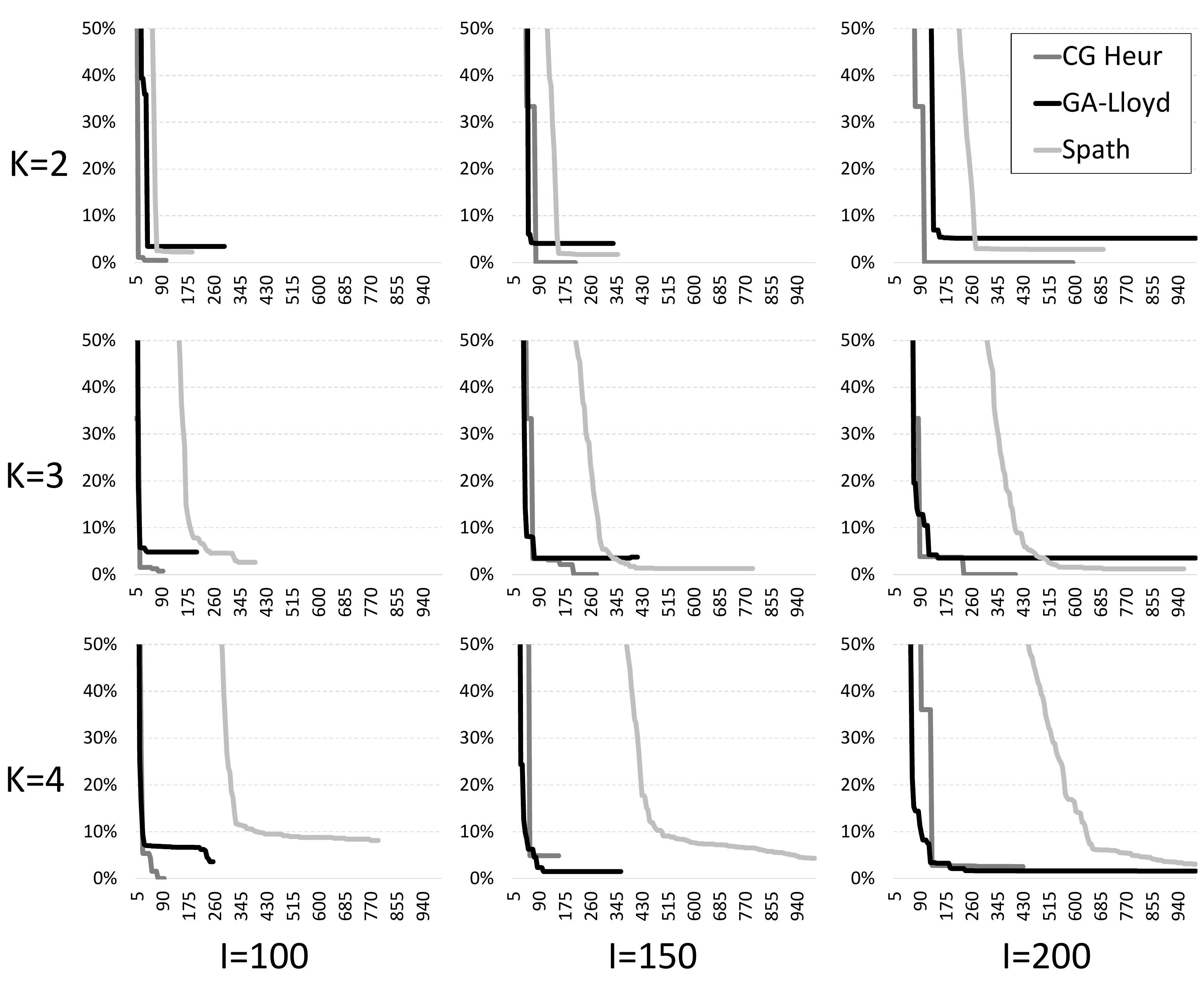}\\
  \caption{Gap from the best objective over time for various K and I}
  \label{Fig:over_time}
\end{figure}

In Figure \ref{Fig:over_time}, we plot the relative gap over time from the best solution obtained in Section 4.2.2. The relative gap at time $t$ of an algorithm is obtained by plugging SSE of the algorithm's current best solution at time $t$ in the first term of the denominator of \eqref{gap_formula}. We plot the result for all 9 pairs of $K = \{2,3,4\}$ and $I = \{100,150,200\}$ in a 3 by 3 grid, where each subplot's horizontal and vertical axes are time (in seconds) and relative gap, respectively. In each plot, dark gray, black, and light gray lines correspond to the CG Heuristic, GA-Lloyd, and Sp{\"a}th algorithms, respectively. The result is based on three instances of each $(K,I)$ pair and each line stops at the average execution time of the corresponding algorithm.

Overall, the CG Heuristic and GA-Lloyd algorithms converge faster than the Sp{\"a}th algorithm. Between the CG Heuristic and GA-Lloyd, it is not easy to decide which algorithm converges faster. When $K=2$, the CG Heuristic converges and terminates faster than GA-Lloyd. However, as $K$ and $I$ increase, there is no trend or significant difference in convergence between the CG Heuristic and GA-Lloyd, although GA-Lloyd terminates later. As $K$ increases, the performance of the Sp{\"a}th algorithm gets worse, although it performs better than GA-Lloyd at termination when $K=2$.

\subsubsection{Time Study of the Column Generation Algorithm}

\begin{figure}[ht]
     \begin{center}
        \subfigure[Type 1]{%
           \includegraphics[scale=0.6]{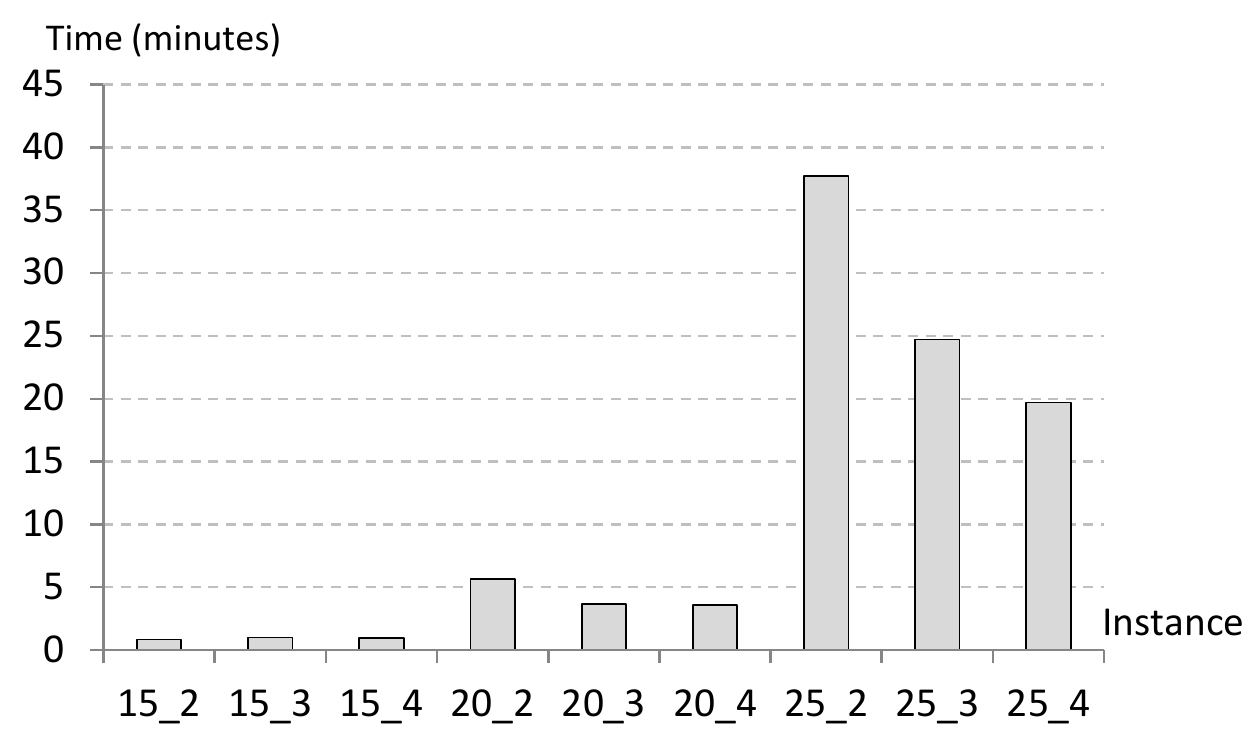}
        }
        \subfigure[Type 2]{
            \includegraphics[scale=0.6]{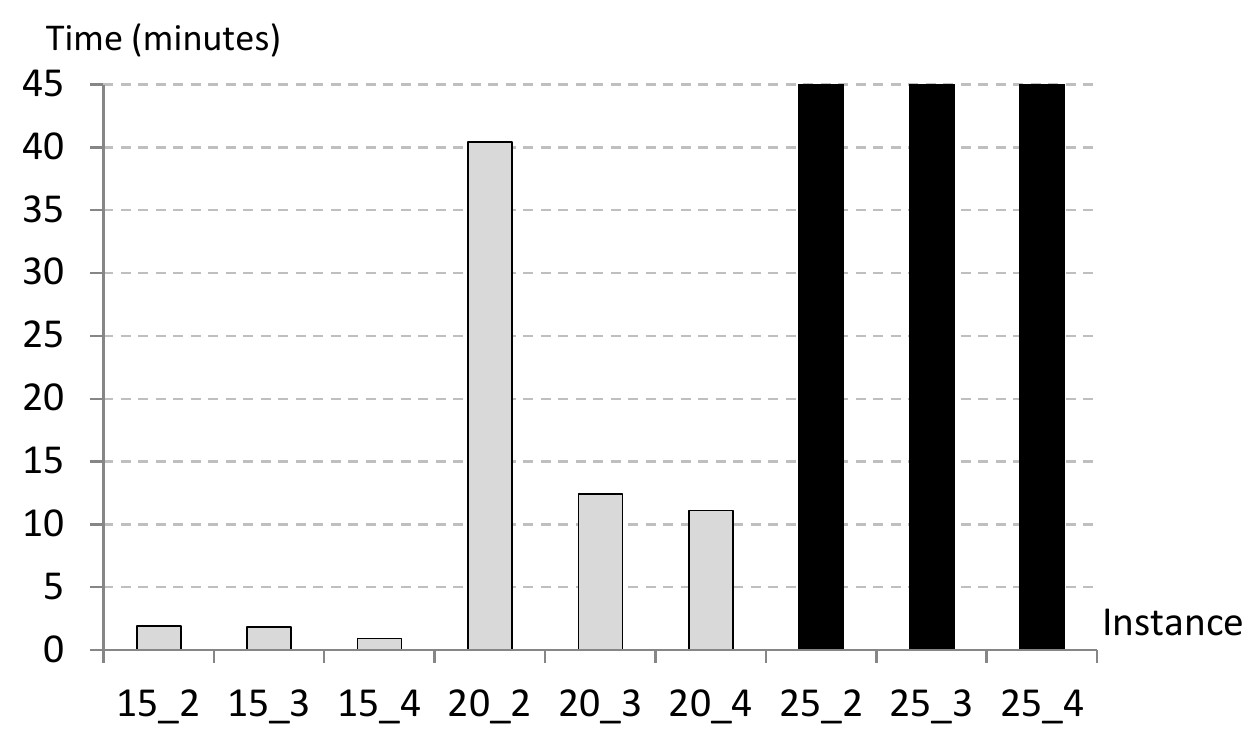}
        }
    \end{center}
    \caption{Column Generation Running Time for Synthetic Data}
    \label{Fig:cgtime_synthetic}
\end{figure}

We present the running time of the CG algorithm for the synthetic data. The running times are of a different magnitude from the one for the real-world data. This is because (1) the computational environments are different and (2) the data has smaller $L$ and the number of attributes $J$. In Figure \ref{Fig:cgtime_synthetic}, we plot the running time of CG. Recall that we did not run CG algorithm for type 2 data with $I = 25$ due to excessive running time. We were not able to get an optimal solution in 10 hours for some of the instances omitted. We observe that the running time drastically increases as $I$ increases. The running time decreases in $K$ with fixed $I$.

\section{Conclusions}\label{Sec:Conclusions}
We propose an exact column generation algorithm, the CG Heuristic algorithm, the GA-Lloyd metaheuristic, and the two-stage algorithm for the resolution of the generalized cluster-wise linear regression problem. We examine the performance of our algorithms on the SKU clustering problem according to seasonal effects using a real-world retail data set from a large retail chain. We find that the column generation exact algorithm can solve small instances to optimality. We use the column generation exact algorithm to benchmark the performance of the GA-Lloyd algorithm and the CG Heuristic algorithm, although the CG algorithm cannot scale well to instances of large sizes. The two-stage algorithm can produce SKU clusters very fast, but with higher objective values than the GA-Lloyd algorithm. The CG Heuristic algorithm performs slightly better than the GA-Lloyd algorithm for some instances, but with much longer running time. The GA-Lloyd algorithm provides a good balance between solution quality and time, and generates SKU clusters with distinctive seasonal patterns efficiently and effectively.
%
%In the future work, we want to solve larger instances of the SKU clustering problem with our column generation exact algorithm. The pricing problem, which we have shown to be NP-hard, is the bottleneck of the column generation algorithm. We believe improved tighter formulations and customized heuristics that exploit problem structures are the key elements for the success of solving the pricing problem efficiently. One added complexity for the larger instances involves the possibility of fractional solutions when solving the linear relaxation of the master problem. We observe fractional solutions in our numerical experiments during the solution process of larger instances using our current column generation algorithm. Therefore, one also needs to consider appropriate branching rules when trying to tackle large sized problems.

\textbf{Acknowledgement:} We are much obliged to Mr. Molham Aref, the CEO of Predictix Inc, for his assistance during the project. Mr. Aref allowed us to use the data and he approved a summer internship by Yan Jiang.

%\bibliography{gCLR}

\bibliographystyle{abbrv}

\appendix
\section*{APPENDIX}
\label{sec_appendix}

%\noindent\textit{Proof of Theorem \ref{Theorem:CLR}:}\\
%Consider an instance of the MSSC problem. There are $I$ entities, and each entity $i$  has an associated vector $\textbf{y}_i=(y_{i1},y_{i2},�,y_{im})$. We want to partition these entities into two clusters. Let the vector $\textbf{y}_i$ be the observations of the dependent variable for entity $i$. In addition, let an identity matrix of size $m$ be the observations of the independent variables  $\textbf{x}_i$, which means
%\begin{equation*}
%\textbf{y}_i =\begin{bmatrix}
%y_{i1}  \\
%y_{i2}  \\
%\vdots  \\
%y_{im}
%\end{bmatrix},\qquad
%\textbf{x}_i = \begin{bmatrix}
%1 & 0 & \ldots
%& 0 \\
%0 & 1 & \ldots
%& 0 \\
%\vdots & \vdots & \ddots
%& \vdots \\
%0 & 0 & \ldots
%& 1
%\end{bmatrix}_{m \times m}
%\end{equation*}
%This transformation can be performed in polynomial time, in the order of  $I \times m^2$. Finding a partition of two clusters with the minimum sum of squared residuals for the CLR problem is equivalent to finding the optimal partition of two clusters for the MSSC problem. For cluster $C_k$ with $k$ being either one or two, the regression coefficients $(\beta_{kj})_{j=1}^m$ compose its centroid with $$\beta_{kj} = \dfrac{\sum_{i \in C_k} y_{ij}}{|C_k|},$$ where $|C_k|$ denotes the cardinality of cluster $C_k$.  Any instance of the MSSC problem can be polynomially reduced to an instance of the CLR problem. Therefore, the CLR problem is NP-hard in a general dimension when the given number of clusters to divide into is two.\hfill$\Box$ \bigskip

\section{Proof of Theorems}
\label{appendix_proof_theorems}

\subsection{Proof of Theorem \ref{Theorem:CLR}}

Intuitively, the generalized CLR problem resembles the MSSC problem, which is known to be NP-hard. We conduct a polynomial transformation from MSSC to a special case of the CLR problem as follows. 

Consider an instance of the MSSC problem with $I$ entities. Each entity $i$  has an associated vector $\textbf{y}_i=(y_{i1},y_{i2},\cdots,y_{iL})$. Let vector $\textbf{y}_i$ be the observations of the dependent variable for entity $i$, and let an identity matrix of size $L$ be the observations of independent variables  $\textbf{x}_i$, which means
\begin{equation}\label{Eq:XMatrix}
\textbf{x}_i = \begin{bmatrix}
1 & 0 & \ldots
& 0 \\
0 & 1 & \ldots
& 0 \\
\vdots & \vdots & \ddots
& \vdots \\
0 & 0 & \ldots
& 1
\end{bmatrix}.
\end{equation}
This yields an instance of the generalized CLR problem of dimension $L \times L$. The regression coefficient $\beta_k$ for cluster $k$ is the centroid of the entities assigned to cluster $k$ with $$\beta_{kl} = \dfrac{\sum_{i \in C_k} y_{il}}{|C_k|}.$$ This proves NP-hardness of the generalized CLR problem.

\hfill$\Box$ \bigskip

\subsection{Proof of Theorem \ref{TH:PricingNPComplete}}

In this section, we show that the pricing problem is NP-complete.

Let us consider a special case of the pricing problem with the observations of the independent variables $\textbf{x}_i$ being an $L \times L$ identity matrix as in \eqref{Eq:XMatrix}. Then the pricing problem becomes
\begin{align*}
\min_{|S|\geq n, \boldsymbol{\beta}} \sum_{i \in S}\sum_{l=1}^L(y_{il} - \beta_l)^2 - \sum_{i \in S} \pi_i.
\end{align*}
Given any cluster $S$ such that $|S| \geq n$, by equating the first order derivative of the pricing objective function to zero, we obtain the optimal $\boldsymbol{\beta}(S)$ as the centroid of vector $\boldsymbol{y}_i$:
\begin{align*}
\beta_l(S) = \dfrac{\sum_{i \in S} y_{il}}{|S|}.
\end{align*}
The Huygen's theorem states that for a given set $S$ of vectors $\boldsymbol{u}_i = (u_{i1}, u_{i2},...,u_{iL})$, the sum of squared distances to the centroid is equal to the sum of squared distances between these vectors divided by two times the cardinality of the set, which mathematically stated reads
\begin{equation*}
\sum_{i \in S}\sum_{j \in S, j \neq i} ||\boldsymbol{u}_i - \boldsymbol{u}_j||_2^2 = 2|S|\sum_{i \in S} ||\boldsymbol{u}_i - \boldsymbol{\bar{u}}(S)||_2^2
\end{equation*}
where $\bar{u}_l(S) = \dfrac{\sum_{i \in S} u_{il}}{|S|}$ and $\boldsymbol{\bar{u}}(S) = (\bar{u}_1(S),\bar{u}_2(S),...,\bar{u}_L(S))$. Based on Huygen's theorem, this special case of the pricing problem can also be stated as:
\begin{align*}
\min_{|S|\geq n} \sum_{i \in S}\sum_{j \in S, j \neq i}||\boldsymbol{y}_i - \boldsymbol{y}_j||_2^2 - 2|S|\sum_{i \in S} \pi_i.
\end{align*}
By using a transformation from the independent set problem, \cite{GareyJohnson1979}, we show that this special case of the pricing problem with this formulation is NP-complete, which implies the pricing problem is NP-complete. 

Let us now formally prove the theorem. We first introduce the independent set problem (\cite{GareyJohnson1979}), a known NP-complete problem, which is used to prove that the pricing problem is NP-complete.\newline
\emph{Instance}: Graph $G = (V,E)$, and a positive integer $M \leq |V|$;\newline
\emph{Question}: Does $G$ contain an independent set of size $M$, i.e., a subset $V'\subseteq V$ with $|V'| = M$ such that no two vertices in $|V'|$ are joined by an edge in $E$?
%The independent set problem is as follows.\newline
%\emph{Instance}: Graph $G = (V,E)$, and a positive integer $K \leq |V|$;\newline
%\emph{Question}: Does $G$ contain an independent set of size $K$ or more, i.e., a subset $V'\subseteq V$ with $|V'| \geq K$ such that no two vertices in $|V'|$ are joined by an edge in $E$?
%Consider the following slight variation of the independent set problem. Let us refer to it as the independent set of size $M$ problem.\newline

%If we can answer the question "Does G contain an independent set of size $M$ for any $M = K, K+1,..., |V|$ in polynomial time, then we can answer the independent set problem in polynomial time since we only need to answer at most $|V|$ such questions. So the independent set of size $M$ problem is also NP-complete.

We next show, through a polynomial reduction from the independent set of size $M$ problem that a constrained version of the pricing problem, which we refer to as ``the subset of size $M$ problem,'' is NP-complete. The subset of size $M$ problem is as follows.\newline
\emph{Instance}: $n$ vectors $(\boldsymbol{u}^1,\boldsymbol{u}^2,...,\boldsymbol{u}^n)$ of dimension $m$ (i.e., $\boldsymbol{u}^i=(u^i_1,u^i_2,...,u^i_m)$), $n$ real numbers $\pi_i$ for $i \in [n]$, and another real number $K$;\newline
\emph{Question}: Is there a subset $S \subseteq \{1,...,n\}$ of vectors with cardinality $|S|=M$ such that $$\sum_{i \in S} \sum_{j \in S, j\neq i} \parallel \boldsymbol{u}^i-\boldsymbol{u}^j \parallel_2^2 - \sum_{i \in S} \pi_i \leq K?$$
\begin{lemma}\label{Lemma:SubsetOfSizeMProblemNPComplete}
The subset of size $M$ problem is NP-complete.
\end{lemma}

\noindent\textit{Proof:} We show NP-completeness of the subset of size $M$ problem using its relationship with the independent set of size $M$ problem. Consider an instance of the independent set of size $M$ problem with graph $G = (V,E)$. To each node $i \in V$, we assign a vector $\boldsymbol{u}^i$ of size $|E|$. For $j = 1,...,|E|$, we have
\begin{equation*}
u_j^i =\left\{
         \begin{array}{ll}
           1, & \hbox{if edge $(i,j) \in E$ and $i < j$;} \\
           -1, & \hbox{if edge $(i,j) \in E$ and $i \geq j$;} \\
           0, & \hbox{otherwise.}
         \end{array}
       \right.
\end{equation*}

Let $k_i$ be the degree of node $i$. If node $i$ is connected to node $j$ by edge $(i,j) \in E$, then
\belowdisplayskip=0pt
\begin{align*}
||\boldsymbol{u}^i - \boldsymbol{u}^j||_2^2 & = (k_i - 1) + (k_j - 1) + (1 - (-1))^2\\
& = k_i + k_j + 2,
\end{align*}
and otherwise,
\begin{equation*}
||\boldsymbol{u}^i - \boldsymbol{u}^j||_2^2 = k_i + k_j.
\end{equation*}

Let $\pi_i = 2(M-1)k_i$ and $K = 0$. We next show that to answer the question whether there is a subset $V'\subseteq V$ with $|V'| = M$ such that $\sum_{i \in V'}\sum_{j \in V', j \neq i} ||\boldsymbol{u}^i - \boldsymbol{u}^j||_2^2 - \sum_{i \in V'} \pi_i \leq 0$ is equivalent to answering the question whether there is an independent subset of size $M$.

If there is an independent subset $V' \subseteq V$ with $|V'| = M$, then $\sum_{i \in V'}\sum_{j \in V', j \neq i} ||\boldsymbol{u}^i - \boldsymbol{u}^j||_2^2- \sum_{i \in V'} \pi_i = \sum_{i \in V'}2(M-1)k_i - \sum_{i \in V'}2(M-1)k_i \leq 0$. If there does not exist an independent subset $V' \subseteq V$ with $|V'| = M$, then $\sum_{i \in V'}\sum_{j \in V', j \neq i} ||\boldsymbol{u}^i - \boldsymbol{u}^j||_2^2 - \sum_{i \in V'} \pi_i\geq \sum_{i \in V'}2(M-1)k_i + 2 \cdot 2 - \sum_{i \in V'}2(M-1)k_i > 0$. Here the first inequality is because there are at least two nodes that are connected by an edge belonging to the subset $V'$.\hfill$\Box$ \bigskip

We now show NP-completeness of the pricing problem by polynomially transforming the subset of size $M$ problem to this problem.
The decision version of our pricing problem is as follows.\\
\emph{Instance}: $I$ vectors $(\boldsymbol{u}^1,\boldsymbol{u}^2,...,\boldsymbol{u}^I)$ of dimension $L$ (i.e., $\boldsymbol{u}^i=(u^i_1,u^i_2,...,u^i_L)$), $I$ real numbers $\pi_i$ for $i \in [I]$, a positive integer $n \leq I$, and another real number $K$.\\
\emph{Question}: Is there a subset $S \subset \{1 ,..., I\}$ such that $\sum_{i \in S}\sum_{j \in S, j\neq i} ||\boldsymbol{u}_i - \boldsymbol{u}_j||_2^2 - \sum_{i \in S} 2|S|\pi_i \leq K$?

If we can answer the pricing problem under the additional constraint that $|S| = M$ for $M = n,...,I$ in polynomial time, then we can answer the original pricing problem in polynomial time. The pricing problem with the additional constraint $|S| = M$ is the subset of size $M$ problem with the same vectors $\boldsymbol{u}^i$ and $I$ real numbers $2|S|\pi_i$. Since the subset of size $M$ problem is NP-complete, so is the pricing problem.
\hfill$\Box$ \bigskip

\section{Random Instance Generation}
\label{appendix_instance_generation}

\subsection{Type 1 Synthetic Data}
\label{appendix_instance_generation_1}
\begin{figure}[ht]
     \begin{center}
        \subfigure[7 patterns]{
            \includegraphics[width=0.4\textwidth]{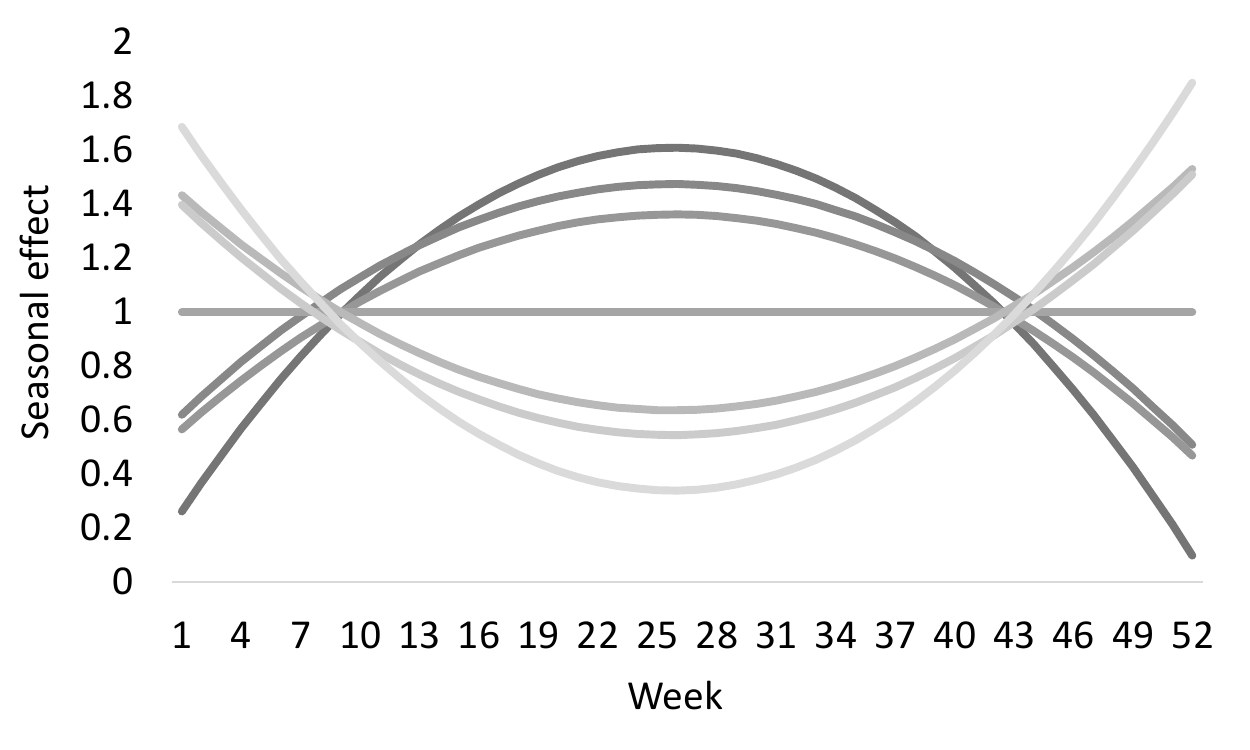} \label{fg_data_gen_a}
        }
        \qquad
        \subfigure[Example of generated instance with 15 entities]{%
           \includegraphics[width=0.4\textwidth]{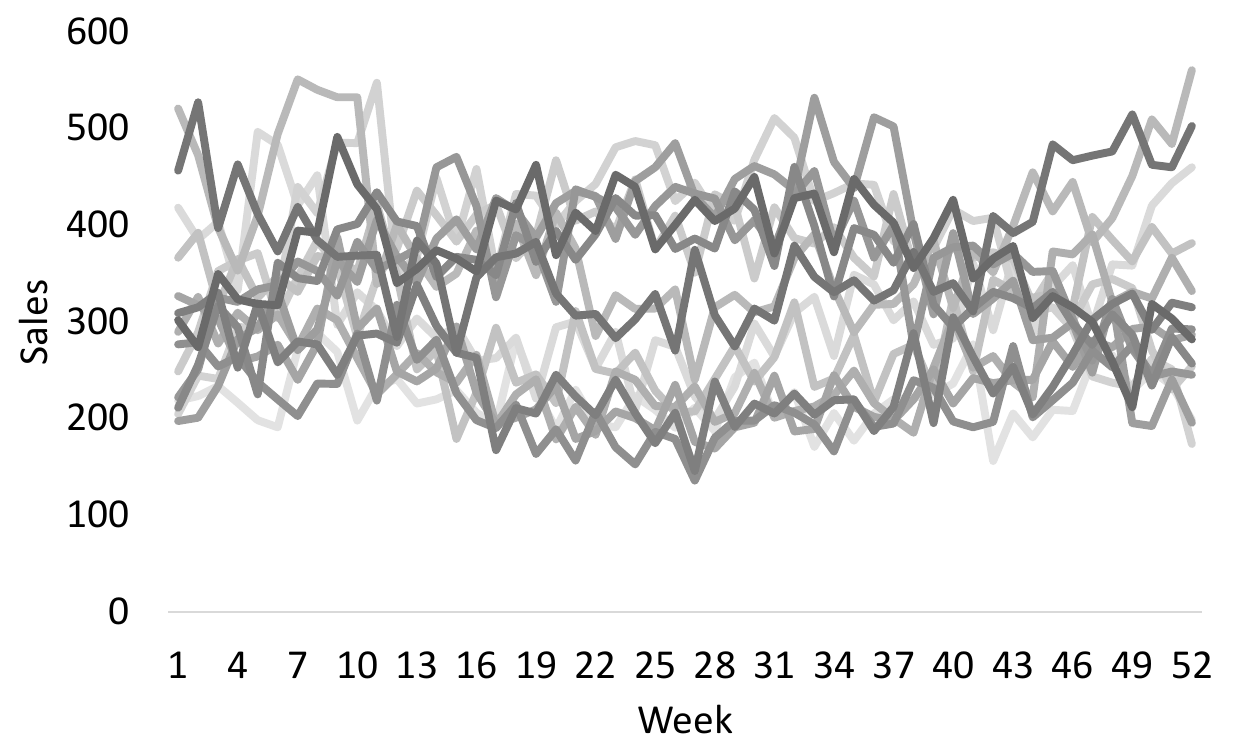} \label{fg_data_gen_b}
        }
    \end{center}
\caption{Seasonal effect patterns and simulated patters} \label{fg_data_gen}
\end{figure}

We generate random instances for the SKU clustering problem based on the following regression model.
\begin{center}
\begin{tabular}{lll}
weekly sales &= &regular and promotional sales + seasonal sales + random noise\\
	& = &discount * $\beta_{\mbox{\scriptsize{discount}}}$ + $t \cdot \beta_t$ + $\varepsilon$,
\end{tabular}
\end{center}
where $t$ is the week index, $\beta_t$ is the regression coefficient for week $t$, and $\beta_{\mbox{\scriptsize{discount}}}$ is the regression coefficient for discount. In order to replicate realistic seasonal effects, we use seven equations for the coefficients of the seasonal effect, motivated from the result in Section 4.2. In Figure \ref{fg_data_gen_a}, the seven equations for the seasonal effect are presented. The horizontal axis represents the week number and the vertical axis represents the ratio between regular / promotional sales and average sales. 

In this section, we denote a uniform random number with lower bound $lb$ and upper bound $ub$ as $U(lb,ub)$. We denote a normal random number with mean $avg$ and standard deviation $std$ as $N(avg,std)$.

We present the overall instance generation algorithm in Algorithm \ref{algo1}. We generate 52 (weeks) time series data points for each identity $i$. The attributes of the final data set includes entity index ($i$), week number ($t$), weekly sales in $t$, and percentage of discount.

In Algorithm \ref{algo1}, we are given $I$ and fixed $L = 52$. For each $i \in \{1,\cdots,I\}$, we generate weekly sales and percentage of discount for 52 weeks based on the following steps. In Line 2, we first randomly generate average sales volume $S_A$ using uniform distribution between 100 and 200. Then in Line 3, we randomly pick a seasonal function from the seven equations illustrated in Figure \ref{fg_data_gen_a}. Next in Line 4, $U(3,6)$ weeks are randomly picked to be the weeks with promotion, where the remaining not selected weeks are without promotion. The discount attribute is generated for promotional weeks by randomly picking among 15\%, 20\%, 25\%, and 30\%. In Lines 5-11, for each $t \in [L]$, we generate sales by summing promotional and seasonal sales, and a random error specified below. In detail, in Lines 6-7, we generate regular and promotional sales. For weeks with promotion, the regular and promotional sales are $S_A \cdot (1 + p_{promo})$. If discount is $15\%$, then $p_{promo}$ is $U(0.4,0.5)$. If discount is 20\%, 25\%, and 30\%, then $p_{promo}$ is drawn from $U(0.5,0.6),U(0.6,0.7)$, and $U(0.7,0.8)$, respectively. For weeks without promotion, $p_{promo} = 0$ and regular and promotional sales are $S_A$. In Lines 8, seasonal sales $D_s$ is obtained by multiplying $f_S(t)$ and $S_A$. In Line 9, random error $\varepsilon$ is generated from normal distribution with zero mean and deviation $S_A/5$. Finally in Line 10, weekly sales in $t$ are generated by summing $D_p$, $D_s$, and $\varepsilon$. In Figure \ref{fg_data_gen_b}, we plot example sales records for 15 entities.

\begin{algorithm}[ht]
\caption{Data generation (type 1)}
\label{algo1}
\begin{algorithmic}[1]   
\vspace{0.1cm}
\REQUIRE $I$ (number of entities or SKUs), $L = 52$
\STATE \textbf{for} $i = 1, \cdots, I$
\STATE \quad Generate average demand $S_A \sim U(100,200)$
\STATE \quad Randomly pick seasonal function $f_S(t)$
\STATE \quad Randomly pick $U(3,6)$ promotional weeks and generate discount attribute
\STATE \quad \textbf{for}  $t = 1, \cdots,L$
\STATE \qquad \textbf{if} $t$ is promotional week, generate $D_p = S_A \cdot (1 + p_{promo})$
\STATE \qquad \textbf{else} generate $D_p = S_A$
\STATE \qquad Calculate $D_s = S_A \cdot f_S(t)$
\STATE \qquad Generate $\varepsilon \sim N(0,\frac{S_A}{5})$
\STATE \qquad $y_{it} = D_p + D_s + \varepsilon$
\STATE \quad  \textbf{end for} 
\STATE \textbf{end for}
\end{algorithmic}
\end{algorithm}

\subsection{Type 2 Synthetic Data}
\label{appendix_instance_generation_2}
In this section, we present a different instance generation procedure based on Algorithm \ref{algo1}. The difference is that we are given the target number of clusters and each entity is randomly assigned to a cluster. For this reason, we have a target solution for each instance where the solution is expected to have a high quality objective function value. 

The overall framework is presented in Algorithm \ref{algo2}. The main difference from Algorithm \ref{algo1} is Line 1. Instead of picking a seasonal function for each entity, we pick a seasonal function for each cluster so that all entities in one cluster can share the seasonal function. Then in Line 3, each entity is assigned to a cluster. Line 4 is identical to Lines 2-11 of Algorithm \ref{algo1} except that the seasonal function is already given by the cluster that $i$ is assigned to in Line 1.

\begin{algorithm}[ht]
\caption{Data generation (type 2)}
\label{algo2}
\begin{algorithmic}[1]   
\vspace{0.1cm}
\REQUIRE $K$ (target number of clusters), $I$ (number of entities or SKUs), $L = 52$
\STATE Create $K$ clusters and randomly pick seasonal function $f_S(t)$ for each cluster.
\STATE \textbf{for} $i = 1, \cdots, I$
\STATE \quad Assign entity $i$ to one of $K$ clusters from Line 1
\STATE \quad Execute Lines 2-11 in Algorithm \ref{algo1} except Line 3
\STATE \textbf{end for}
\end{algorithmic}
\end{algorithm}

\begin{figure}[ht]
\center
  \includegraphics[scale=0.7]{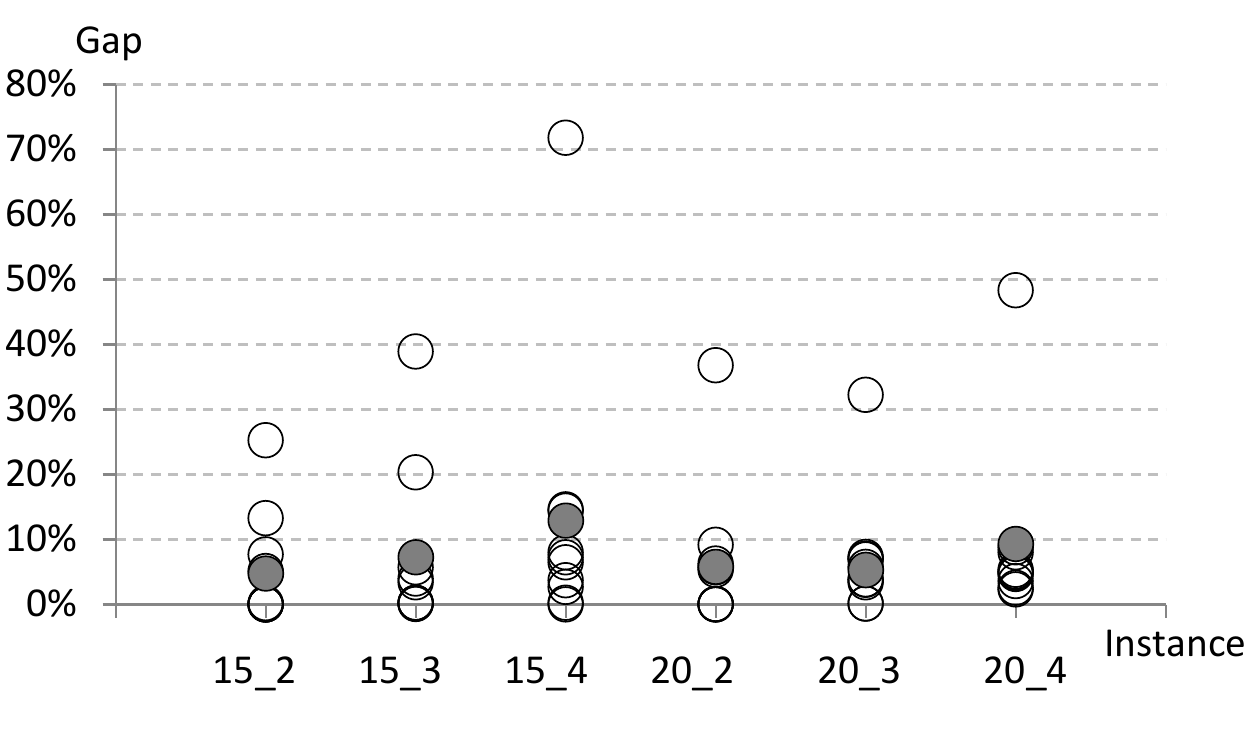}\\
  \caption{Relative gap between optimal and target solution}
  \label{Fig:opt_vs_true}
\end{figure}

Because we generate instances with a target solution, we evaluate the difference between the target solution and optimal solution. In Figure \ref{Fig:opt_vs_true}, we present the gap between the target and optimal solution. Each circle represents the different instances and we plot for 10 instances for each $I\_K$. We observe that the target solutions are within the 10\% gap from the optimal solutions for 51 instances, where 9 instances have gaps greater than 10\%. We also present the average gaps in Table \ref{tab:gaps_target_opt}. 

\begin{table}
\centering
\small
\begin{tabular}{|l|l|l|l|}
\hline
 &	K=2 &	K=3 &	K=4 \\ \hline
I=15 &	5.1\%	& 7.3\% &	12.9\% \\ \hline
I=20 &	6.3\%	& 6.9\%	& 9.3\% \\ \hline
\end{tabular}
\caption{Relative gap between optimal and target solution} \label{tab:gaps_target_opt}
\end{table}

This gives a justification to evaluate algorithms for larger instances based on the gap from the target solution. However, in our experiment for larger instances, we observed that most of the proposed algorithms give a better objective function value than the target solution. Hence, we did not use the target solution to evaluate the algorithms in the paper. However, the target solutions are available on the website stated in Section \ref{Sec:Experiments}.

\section{Restricted Master Problem}
\label{appendix_Restricted_Master_Problem}
\vspace{-0.2cm}

In this section, we present a stabilized version of the master problem \eqref{Eq:MasterProblem} - \eqref{constraints:partition:assignment}, referred to as restricted master problem, by applying the technique of du Merle \textit{et al.} \cite{duMerle1999}. For iteration $k$, the restricted master problem is written as
\begin{align}
\min \quad & \sum_{S\in \mathscr{S}} c_Sz_S - \boldsymbol{\delta}^{(k)} \boldsymbol{q}^- + \boldsymbol{\delta}^{(k)} \boldsymbol{q}^+ \label{Eq:RestrictedMasterProblem}\\
&\sum_{S\in \mathscr{S}} z_S = K \label{constraints:Restrictedpartition:group}\\
&\sum_{S\in \mathscr{S}} a_{iS}z_S  - \boldsymbol{q}^- + \boldsymbol{q}^+ = 1 \quad & i \in [I] \label{constraints:Restrictedpartition:assignment}\\
&0 \leq \boldsymbol{q}^- \leq \boldsymbol{\xi}^{(k)}\\
&0 \leq \boldsymbol{q}^+ \leq \boldsymbol{\xi}^{(k)} \label{restr222}\\ 
&z_S \in \{0,1\}\quad &S \in \mathscr{S},\nonumber
\end{align}
\vspace{0.2cm}

\noindent which is obtained by introducing perturbation variables $\boldsymbol{q}^-$ and $\boldsymbol{q}^+$ and stabilization parameters $\boldsymbol{\delta}^{(k)}$ and $\boldsymbol{\xi}^{(k)}$.

\end{document}